\documentclass[11pt]{article}

\usepackage[preprint]{acl}

\usepackage{times}
\usepackage{latexsym}
\usepackage[T1]{fontenc}
\usepackage[utf8]{inputenc}
\usepackage{microtype}
\usepackage{inconsolata}
\usepackage[most]{tcolorbox}
\usepackage{makecell}
\usepackage{amsmath, amssymb, amsfonts}
\usepackage{booktabs}
\usepackage{multirow}
\usepackage{graphicx}
\usepackage{xcolor}
\usepackage{url}
\usepackage{enumitem}
\usepackage{algorithm}
\usepackage{algorithmic}
\usepackage{mathtools}
\usepackage{hyperref}
\usepackage[table]{xcolor}
\title{SAGE: Answer-Conditioned Uncertainty Targets for Verbal Uncertainty Alignment}

\author{
Kaiwen Shi\thanks{Equal contribution.} \quad
Zheyuan Zhang\footnotemark[1] \quad
Yanfang Ye\thanks{Corresponding author.} \\
University of Notre Dame \\
\texttt{\{kshi3,yye7\}@nd.edu} 
}

\begin{document}
\maketitle

\begin{abstract}
Large language models increasingly express uncertainty through natural-language statements, yet these expressions often fail to reflect the model's sampled behavior. We study verbal uncertainty alignment as a distributional calibration problem: the appropriate uncertainty target for a prompt should be estimated from repeated model outputs rather than from an isolated response. However, group rollouts alone are insufficient, since the resulting target must provide a useful training signal. Existing targets only partially satisfy this requirement. Therefore, we propose \textbf{SAGE} (\textbf{S}emantic-\textbf{A}nswer \textbf{G}uided \textbf{E}ntropy), a group-level uncertainty target that constructs an answer-conditioned uncertainty geometry over sampled responses. SAGE preserves categorical, numeric, and symbolic answer distinctions while maintaining a smooth and scale-preserving calibration signal. We further apply this target through \textbf{Group-Uncertainty Preference Optimization} (GUPO), an uncertainty-channel training framework that supervises verbal uncertainty expressions rather than the full response. Experiments across factual, mathematical, and multiple-choice reasoning tasks show improved uncertainty ranking, lower calibration error, and reduced overconfidence. Our code is available \href{https://anonymous.4open.science/r/SAGE-6017/}{here}.
\end{abstract}

\section{Introduction}
\begin{figure*}[t]
    \centering
    \includegraphics[width=\linewidth]{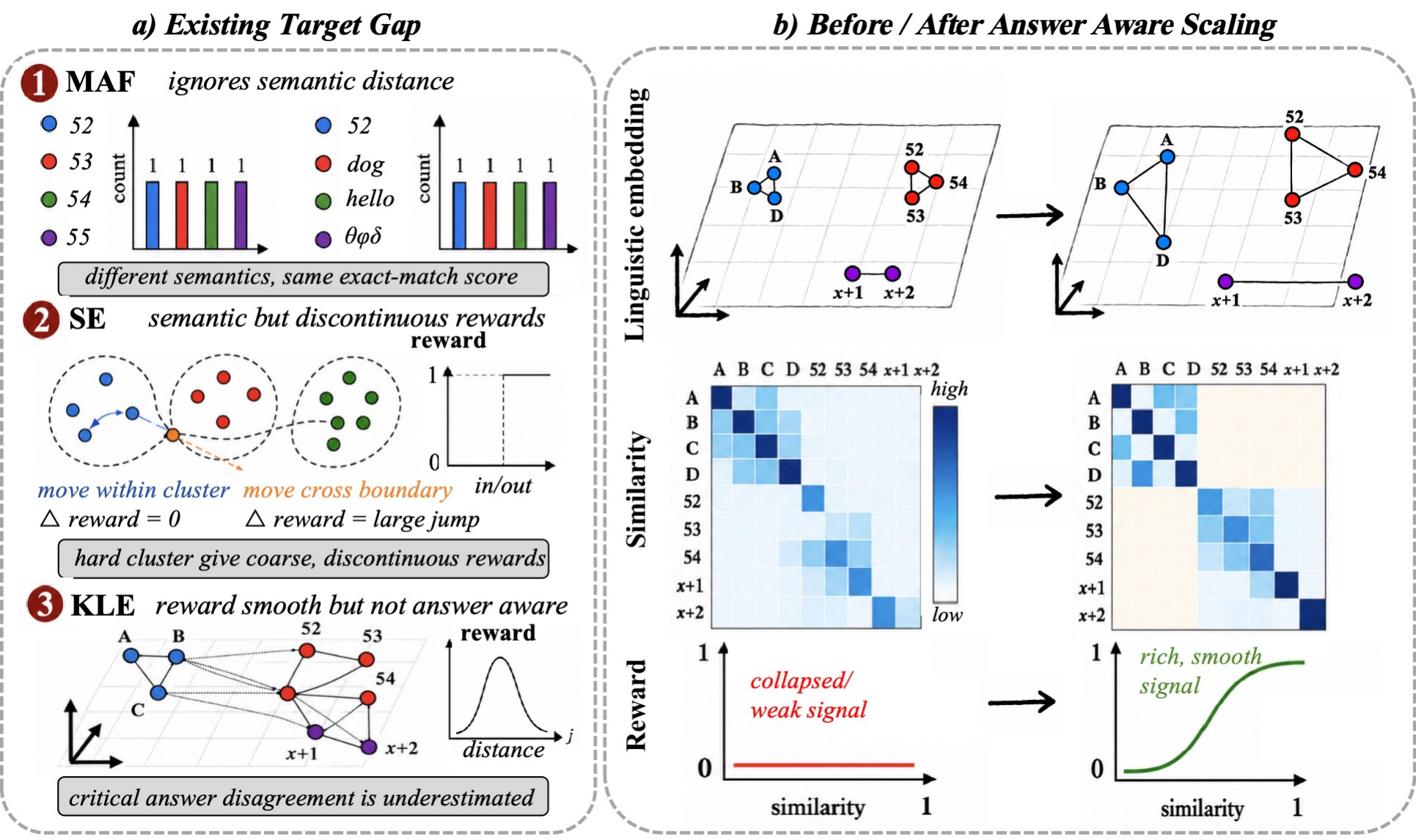}
    \caption{
    \textbf{Our motivation and key idea.} Existing group-level targets each miss a key requirement for verbal uncertainty calibration: MAF is answer-faithful but coarse, SE captures semantic equivalence but produces discontinuous rewards, and KLE is smooth but can underestimate task-critical answer disagreement in generic embedding space. SAGE addresses these limitations by injecting answer-aware structure into the response geometry while preserving smooth kernel-based entropy. This expands task-critical differences within compressed semantic neighborhoods and yields a smoother, more answer-faithful, and scale-preserving uncertainty target, providing a stronger reward signal for verbal uncertainty alignment.
    }
    \label{fig:motivation}
\end{figure*}
Recent progress in large language models (LLMs) \cite {ye2025llms4all,chen2025clear,chen2025obvious} has achieved remarkable success and expanded their use in open-ended generation tasks \cite{Meng2024TheAO, Belkhouribchia2025LargeLM}. However, this success is accompanied by serious reliability challenges: model outputs can appear fluent and plausible even when the underlying answer is uncertain or unsupported \cite{Farquhar2024DetectingHI}. In real-world scenarios, hidden uncertainty may lead to hallucinations with severe consequences. These failures are not simply matters of imperfect text generation; they can mislead users, amplify unsafe decisions, and, in extreme cases, cause physical harm \cite{Savage2024LargeLM,Omar2025MultimodelAA}.

To mitigate such risks, researchers have explored different ways to represent model uncertainty. Internal signals, such as token probabilities \cite{Gupta2024LanguageMC} or hidden states \cite{Azaria2023TheIS}, are informative but require model access and are difficult to interpret. Sampling-based signals reveal behavioral instability across repeated generations \cite{Kuhn2023SemanticUL, Nikitin2024KernelLE}, but they are costly. Meanwhile, asking an LLM to verbalize its own uncertainty provides a direct and practical signal \cite{Xiong2023CanLE, Lin2022TeachingMT}. A statement such as ``I am uncertain'' or ``I am 80\% confident'' is immediately visible in the model output, available even in black-box settings, and easy for users or downstream applications to interpret. However, this convenience is useful only if the expressed uncertainty matches the reliability of the model's actual behavior, i.e., if the verbal uncertainty is calibrated \cite{Kapoor2024LargeLM}. Without calibration, verbal uncertainty becomes unreliable and can turn model errors into misplaced trust.

A key challenge is constructing the target used to align verbal uncertainty. Since uncertainty should reflect the model's behavioral uncertainty rather than the surface form of a single answer, the central question is what target the model should learn \cite{Geng2023ASO}. Existing targets capture different aspects of uncertainty, but each leaves an important gap. Maximum answer frequency (MAF) is answer-relevant but ignores semantic distance \cite{Cole2023SelectivelyAA}. Semantic Entropy  (SE) handles paraphrases through meaning-equivalence clusters, but hard clustering makes the signal coarse and discontinuous \cite{Kuhn2023SemanticUL}. Kernel language entropy (KLE) is smoother, but generic embedding geometry can blur task-critical answer distinctions, such as option labels, numeric values, or symbolic expressions \cite{Nikitin2024KernelLE}. Thus, target quality becomes an optimization bottleneck: \textbf{without a reliable uncertainty target, the reward signal can collapse into noise or shortcuts,} leaving the optimizer unable to distinguish calibrated uncertainty from superficial response patterns.

To address this target gap, we propose \textbf{SAGE} (\textbf{S}emantic-\textbf{A}nswer \textbf{G}uided \textbf{E}ntropy), a new group-level self-uncertainty target that reformulates verbal uncertainty alignment through answer-conditioned uncertainty geometry. The key idea is that an uncertainty target should not be determined solely by generic semantic variation; it should also reflect whether sampled generations remain compatible under the task-specific answer structure. This allows SAGE to preserve smooth reward variation while correcting a common failure mode of generic semantic metrics, where incompatible option labels, distinct numerical values, or non-equivalent symbolic expressions are treated as near-equivalent. By imposing this answer-conditioned structure, SAGE turns repeated generations into a structured uncertainty signal rather than an unorganized set of samples. Building on this target, we further introduce \textbf{Group-Uncertainty Preference Optimization} (GUPO), which uses SAGE as the supervision signal to align verbal uncertainty with sampled model behavior. Experiments on factual QA, mathematical reasoning, and multiple-choice understanding show that our framework improves uncertainty ranking, reduces calibration error, and mitigates overconfidence across distinct answer formats.

Our contributions are as follows:
\begin{itemize}
    \item We show that verbal uncertainty alignment is a distributional calibration problem, and identify a reward-signal bottleneck that limits existing group-level uncertainty targets.
    \item We introduce \textbf{SAGE}, a semantic-answer guided entropy target, together with \textbf{GUPO}, an uncertainty-channel preference framework that optimizes verbal uncertainty expressions instead of full responses.
    \item We validate the approach across factual, mathematical, and multiple-choice reasoning tasks, where our framework consistently outperforms existing baselines in uncertainty ranking, calibration error, and overconfidence reduction.
\end{itemize}

\section{Problem formulation}

\subsection{Distributional Blindness}

Existing work attempts to address uncertainty calibration through response-level methods, such as supervised fine-tuning \cite{Jang2025VerbalizedCT} and pairwise preference optimization \cite{Zhang2025DirectCA, Zhang2025ReinforcementLF}. However, a single response can reveal what the model sampled, but it cannot reveal whether the underlying probability mass is concentrated on that answer or dispersed across plausible alternatives, thus creating a structural mismatch for self-uncertainty calibration.

To be specific, supervised fine-tuning teaches the model to reproduce a target uncertainty expression by maximizing the likelihood of the supervised output tokens \cite{Lin2022TeachingMT, Chaudhry2024FinetuningLM, Jang2025VerbalizedCT}, instead of directly learning how uncertainty should change with the dispersion of the model's own response distribution. Similarly, pairwise preference optimization establishes only a local gradient by updating the policy to maximize the log-likelihood margin between two static responses \cite{Li2026ORCEOA, Zhang2025DirectCA}, which lack sufficient information to determine whether the policy is sampling from a stable or highly entropic region. As a result, response-level training can adjust the form or scale of verbal uncertainty, but it does not naturally align uncertainty with the model's sampled response distribution.

\begin{figure}[t]
    \centering
    \includegraphics[width=\linewidth]{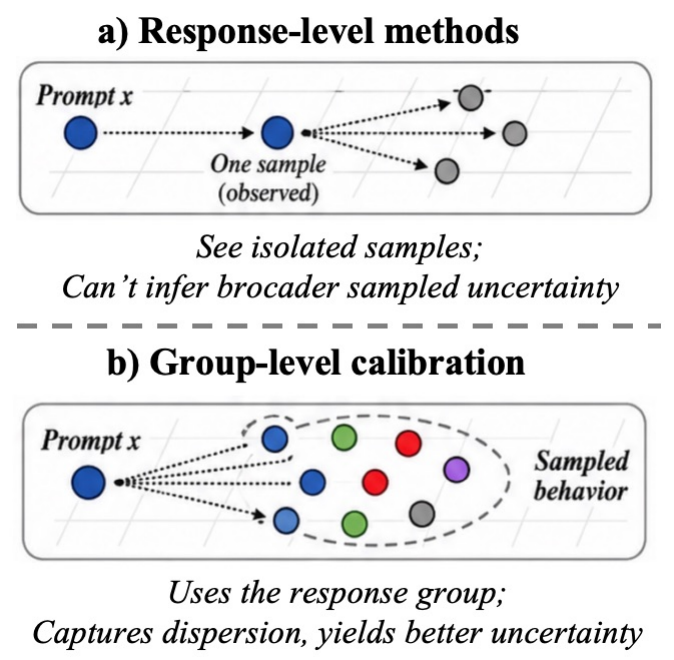}
    \caption{
    Illustration of the distributional gap in verbal uncertainty calibration.
    (a) A single response cannot reveal whether it comes from a stable or unstable response distribution.
    (b) Distributional calibration uses repeated samples: stable groups justify lower uncertainty, while dispersed groups require higher uncertainty.
    }
    \label{fig:distributional_gap}
\end{figure}

\begin{figure*}[t]
    \centering
    \includegraphics[width=\textwidth]{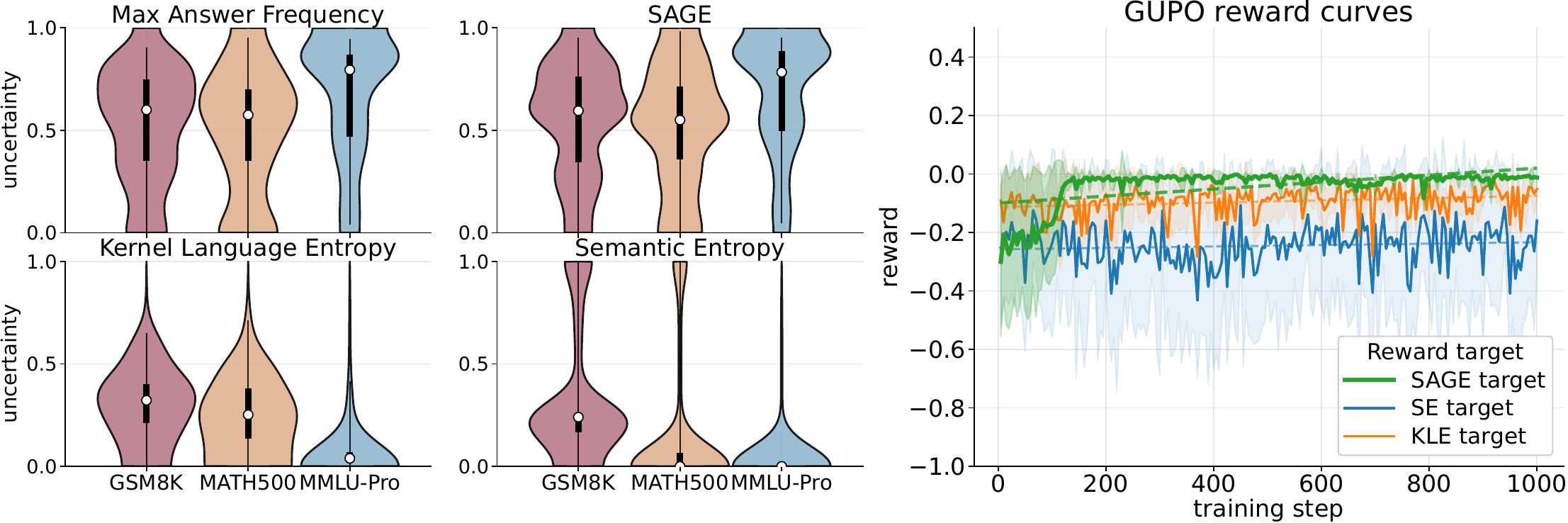}
    \caption{
    Uncertainty distributions produced by different group-level targets on the same 20 sampled responses from the evaluation set. Semantic Entropy (SE) and Kernel Language Entropy (KLE) can both underestimate answer-level uncertainty: SE collapses variation through hard semantic clusters, while KLE smooths over task-critical answer disagreement in embedding space. This motivates a target that is smooth, answer-aware, and scale-preserving.
    }
    \label{fig:motivation_uncertainty_targets}
\end{figure*}

\subsection{Group-Uncertainty Preference Optimization}

After constructing a group-level uncertainty target \(t(G_x)\), we apply the calibration signal to the verbal uncertainty expression rather than to the full response. For a response \(y_i=(z_i,v_i)\), the answer-bearing content \(z_i\) provides context, while the uncertainty expression \(v_i\) is the object of calibration. Let \(\hat u(v_i)\in[0,1]\) denote the uncertainty value expressed by \(v_i\). We define
\[
r_i=-\ell\big(\hat u(v_i),t(G_x)\big),
\]
where larger values indicate that the expressed uncertainty is closer to the group-level uncertainty target.

We use these scores to form a group-relative preference distribution over uncertainty expressions:
\[
p_i^{\mathrm{cal}}
=
\frac{\exp(r_i/T)}
{\sum_{j=1}^{K}\exp(r_j/T)},
\]
where \(T\) controls preference sharpness. GUPO applies this preference to the uncertainty channel,
\[
v_i \mid (x,z_i),
\]
rather than treating the entire response as the optimization target. Conceptually, this corresponds to
\[
\begin{aligned}
\mathcal{L}_{\mathrm{GUPO}}
=
-\mathbb{E}_{x,G_x}
\Bigg[
&\sum_{i=1}^{K}
\operatorname{stopgrad}(p_i^{\mathrm{cal}})
\\
&\cdot
\log \pi_\theta(v_i\mid x,z_i)
\Bigg].
\end{aligned}
\]
In practice, this can be implemented by applying the calibration loss only to the uncertainty-expression span. The main focus of this work is the construction of \(t(G_x)\), which we study next.

\section{Methodology}

\subsection{Why Existing Group Targets Give Weak Rewards}

Group rollouts expose the response distribution, but they do not automatically produce a useful uncertainty target. In group-based optimization, the model only receives a useful signal if the group-level target changes in ways that reflect meaningful uncertainty differences, while a poor target can make repeated sampling ineffective. If the target is too coarse, too discontinuous, or misaligned with answer-level disagreement, then the optimizer receives little guidance about how verbal uncertainty should change.

\paragraph{Weak Group Rewards.}
Figure~\ref{fig:motivation} illustrates why existing group-level targets provide weak reward signals for verbal uncertainty calibration. Given a response group \(G_x=\{y_i\}_{i=1}^K\) and extracted answers \(\{a_i\}_{i=1}^K\), the simplest stability target is maximum answer frequency:
\[
s_{\mathrm{MAF}}(G_x)
=
\max_a
\frac{1}{K}
\sum_{i=1}^{K}
\mathbb{I}[a_i=a].
\]
Since Maximum Answer Frequency \cite{Cole2023SelectivelyAA} measures answer stability, its corresponding uncertainty target can be written as \(t_{\mathrm{MAF}}(G_x)=1-s_{\mathrm{MAF}}(G_x)\). MAF is answer-faithful: if most samples produce the same final answer, the model should generally express lower uncertainty. However, as shown in Fig.~\ref{fig:motivation}(1), MAF only counts exact answer agreement. It maps semantically different response groups to the same score whenever their majority counts are identical:
\[
G_x \neq G_x',
\qquad
s_{\mathrm{MAF}}(G_x)=s_{\mathrm{MAF}}(G_x').
\]
Thus, MAF preserves final-answer agreement but discards semantic distance, reasoning variation, and secondary answer alternatives.

Semantic Entropy (SE) \cite{Kuhn2023SemanticUL} improves on exact counting by clustering responses into meaning-equivalence classes. Let \(c_i\) denote the semantic cluster of response \(y_i\) and \(p(c)\) be the empirical cluster frequency:
\[
t_{\mathrm{SE}}(G_x)
=
-\sum_c p(c)\log p(c).
\]
This handles paraphrases and aliases better than raw answer frequency. Yet, as shown in Fig.~\ref{fig:motivation}(2), SE depends on hard cluster membership. A small movement within the same cluster gives no reward change,
\[
c_i=c_i'
\quad\Rightarrow\quad
\Delta t_{\mathrm{SE}}=0,
\]
while crossing a cluster boundary can cause a discrete jump. Therefore, SE produces coarse and discontinuous rewards: many meaningful changes in similarity or reasoning variation are invisible unless they alter the cluster assignment.

Kernel Language Entropy (KLE) \cite{Nikitin2024KernelLE} replaces hard clusters with continuous similarities:
\[
t_{\mathrm{KLE}}(G_x)
=
H_{\mathrm{vN}}(K),
\qquad
K_{ij}=k(y_i,y_j),
\]
where \(K\) is a semantic similarity kernel over generated responses. This gives a smoother uncertainty signal than SE. However, as shown in Fig.~\ref{fig:motivation}(3), its smoothness comes from generic embedding geometry rather than the task-specific answer geometry. As a result, KLE may assign high similarity to outputs that are linguistically close but answer-wise incompatible:
\[
k(y_i,y_j)\ \text{high}
\quad \not\Rightarrow \quad
a_i \equiv a_j.
\]
This is problematic for structured answer spaces such as multiple-choice labels, numerical answers, or symbolic expressions, where small surface differences can encode decisive answer disagreement.

Together, these limitations show that a useful uncertainty target must be both answer-sensitive and reward-smooth. Otherwise, the reward either collapses into coarse counts, discontinuous cluster changes, or misleading embedding similarity, leaving the optimizer with weak credit assignment for verbal uncertainty calibration.

\paragraph{Empirical Observation.}
Figure~\ref{fig:motivation_uncertainty_targets} reveals a reward-signal bottleneck. Under the same repeated-sampling groups, Maximum Answer Frequency preserves a broad stability range, whereas SE and KLE are compressed toward low uncertainty, thereby underestimating disagreement among sampled answers.

This compression arises because generic semantic similarity may miss task-specific answer disagreement. For example, options A, B, C, and D can be close in embedding space despite being mutually exclusive; nearby numeric answers such as 52, 53, and 54 may appear similar although only one is correct; and symbolic answers such as \(x+1\) and \(x+2\) can be linguistically close while mathematically distinct. Consequently, SE and KLE assign weakly separated uncertainty targets to groups that require different uncertainty levels. Their reward curves therefore improve only marginally and quickly plateau, suggesting limited guidance for adjusting verbal uncertainty. In contrast, SAGE preserves answer-level distinctions while maintaining a smooth signal, yielding a more informative reward trajectory.

\begin{table*}[t]
\centering
\small
\renewcommand{\arraystretch}{1.16}
\setlength{\tabcolsep}{5.4pt}
\definecolor{oursblue}{RGB}{226,235,248}
\definecolor{deltagrey}{RGB}{238,238,238}
\definecolor{headergrey}{RGB}{245,245,245}

\newcommand{\best}[1]{\textbf{\underline{#1}}}
\newcommand{\oc}{\cellcolor{oursblue}}
\newcommand{\dc}{\cellcolor{deltagrey}}

\begin{tabular}{llcccccc}
\toprule
\rowcolor{headergrey}
\textbf{Benchmark}
& \textbf{Method}
& \multicolumn{3}{c}{\textbf{Calibration}}
& \multicolumn{3}{c}{\textbf{High-Confidence Accuracy}} \\
\cmidrule(lr){3-5}
\cmidrule(lr){6-8}
\rowcolor{headergrey}
&
& \textbf{Brier $\downarrow$}
& \textbf{ECE $\downarrow$}
& \textbf{Spear. $\uparrow$}
& \textbf{@80 $\uparrow$}
& \textbf{@60 $\uparrow$}
& \textbf{@50 $\uparrow$} \\
\midrule

\multirow{10}{*}{MMLU-Pro}
& Direct Verbalized & 0.126 & 0.246 & 0.069 & 0.303 & 0.312 & 0.325 \\
& Direct Verbalized + DCA & 0.113 & 0.222 & 0.124 & 0.356 & 0.412 & 0.445 \\
& CoT & 0.409 & 0.590 & 0.299 & 0.291 & 0.296 & 0.340 \\
& CoT + DCA & 0.094 & 0.179 & 0.100 & 0.319 & 0.342 & 0.355 \\
& CSFT & 0.045 & 0.120 & 0.195 & 0.347 & 0.388 & 0.380 \\
& LoVeC-DPO & 0.051 & 0.144 & 0.261 & 0.403 & 0.438 & 0.460 \\
\cmidrule(lr){2-8}
& \oc GUPO + Semantic Entropy
& \oc 0.214 & \oc 0.316 & \oc 0.090 & \oc 0.188 & \oc 0.183 & \oc 0.180 \\
& \oc GUPO + Kernel Language Entropy
& \oc 0.068 & \oc 0.205 & \oc -0.010 & \oc 0.163 & \oc 0.167 & \oc 0.180 \\
& \oc \textbf{GUPO + SAGE}
& \oc \best{0.037} & \oc \best{0.113} & \oc \best{0.572}
& \oc \best{0.412} & \oc \best{0.512} & \oc \best{0.570} \\
& \dc $\Delta$ vs. Best Baseline
& \dc $-$0.008 & \dc $-$0.007 & \dc +0.273
& \dc +0.009 & \dc +0.074 & \dc +0.110 \\
\midrule

\multirow{10}{*}{MATH-500}
& Direct Verbalized & 0.359 & 0.546 & 0.246 & 0.563 & 0.617 & 0.660 \\
& Direct Verbalized + DCA & 0.388 & 0.578 & 0.171 & 0.550 & 0.617 & 0.580 \\
& CoT & 0.396 & 0.582 & 0.175 & 0.588 & 0.617 & 0.600 \\
& CoT + DCA & 0.287 & 0.479 & 0.074 & 0.588 & 0.567 & 0.580 \\
& CSFT & 0.064 & 0.143 & \best{0.500} & \best{0.625} & 0.700 & \best{0.740} \\
& LoVeC-DPO & 0.121 & 0.220 & 0.352 & 0.600 & 0.633 & 0.660 \\
\cmidrule(lr){2-8}
& \oc GUPO + Semantic Entropy
& \oc 0.190 & \oc 0.282 & \oc 0.426 & \oc 0.575 & \oc 0.650 & \oc 0.720 \\
& \oc GUPO + Kernel Language Entropy
& \oc 0.0205 & \oc 0.0348 & \oc 0.451 & \oc 0.588 & \oc 0.700 & \oc 0.720 \\
& \oc \textbf{GUPO + SAGE}
& \oc \best{0.0202} & \oc \best{0.0238} & \oc 0.463
& \oc 0.613 & \oc \best{0.717} & \oc \best{0.740} \\
& \dc $\Delta$ vs. Best Baseline
& \dc $-$0.044 & \dc $-$0.119 & \dc $-$0.037
& \dc $-$0.012 & \dc +0.017 & \dc +0.000 \\
\midrule

\multirow{10}{*}{TriviaQA}
& Direct Verbalized & 0.394 & 0.551 & 0.132 & 0.388 & 0.433 & 0.440 \\
& Direct Verbalized + DCA & 0.235 & 0.402 & 0.507 & 0.450 & 0.517 & 0.570 \\
& CoT & 0.341 & 0.504 & 0.307 & 0.419 & 0.467 & 0.500 \\
& CoT + DCA & 0.209 & 0.368 & 0.523 & 0.469 & 0.508 & 0.560 \\
& CSFT & 0.119 & 0.177 & 0.241 & 0.494 & 0.517 & 0.540 \\
& LoVeC-DPO & 0.107 & 0.118 & 0.015 & 0.481 & 0.467 & 0.480 \\
\cmidrule(lr){2-8}
& \oc GUPO + Semantic Entropy
& \oc 0.153 & \oc 0.247 & \oc 0.404 & \oc 0.480 & \oc 0.530 & \oc 0.540 \\
& \oc GUPO + Kernel Language Entropy
& \oc 0.123 & \oc 0.227 & \oc 0.268 & \oc 0.490 & \oc 0.480 & \oc 0.520 \\
& \oc \textbf{GUPO + SAGE}
& \oc \best{0.055} & \oc \best{0.031} & \oc \best{0.617}
& \oc \best{0.506} & \oc \best{0.583} & \oc \best{0.650} \\
& \dc $\Delta$ vs. Best Baseline
& \dc $-$0.052 & \dc $-$0.087 & \dc +0.094
& \dc +0.012 & \dc +0.066 & \dc +0.080 \\

\bottomrule
\end{tabular}
\caption{
Calibration results across three benchmarks.
Lower values are better for Brier score and ECE, while higher values are better for Spearman correlation and high-confidence subset accuracy.
In our uncertainty framing, high-confidence subset accuracy corresponds to evaluating whether lower-uncertainty predictions are more reliable.
$\Delta$ is computed against the best non-GUPO baseline for each metric.
}
\label{tab:calibration_results_all}
\end{table*}

\subsection{SAGE: Semantic-Answer Guided Entropy}

We propose \textbf{Semantic-Answer Guided Entropy} (SAGE) to construct such a target. SAGE keeps the continuous kernel view of linguistic entropy, but modifies the response geometry using task-specific answer equivalence.

For each response \(y_i=(z_i,v_i)\), let \(e(z_i)\) denote a continuous representation of the answer-bearing content. We define the linguistic kernel as
\[
K_{\mathrm{ling}}(y_i,y_j)
=
\exp\left(
-\frac{\|e(z_i)-e(z_j)\|^2}
{2\sigma_{\mathrm{ling}}^2}
\right),
\]
which measures smooth semantic similarity between responses.

Let \(a_i\) be the extracted final answer from \(z_i\). We first define a task-specific binary answer-equivalence kernel
\[
K_{\mathrm{ans}}^{\mathrm{bin}}(a_i,a_j)\in\{0,1\},
\]
where equivalent answers receive value \(1\). Since a hard kernel may over-separate cross-answer pairs, we use a soft floor:
\[
K_{\mathrm{ans}}^{\mathrm{soft}}(a_i,a_j)
=
c+(1-c)K_{\mathrm{ans}}^{\mathrm{bin}}(a_i,a_j),
c\in[0,1].
\]
Here, \(c=0\) gives hard answer separation, while \(c=1\) removes answer geometry and recovers linguistic KLE.

The final answer-aware kernel is
\[
K_{\mathrm{A}}(y_i,y_j)
=
K_{\mathrm{ling}}(y_i,y_j)^\alpha
\left(
K_{\mathrm{ans}}^{\mathrm{soft}}(a_i,a_j)
\right)^\beta ,
\]
where \(\alpha\) and \(\beta\) control linguistic and answer-aware similarity. Diagonal entries are set to \(1\).

We normalize \(K_{\mathrm{A}}\) into a density matrix:
\[
P_{\mathrm{A}}
=
\frac{K_{\mathrm{A}}}
{\operatorname{tr}(K_{\mathrm{A}})}.
\]
SAGE is the normalized von Neumann entropy:
\[
H_{\mathrm{SAGE}}(G_x)
=
-\frac{1}{\log K}
\operatorname{tr}
\left(
P_{\mathrm{A}}\log P_{\mathrm{A}}
\right).
\]
Lower entropy indicates concentrated responses under linguistic and answer-aware similarity, while higher entropy indicates dispersion across meanings, reasoning paths, or final answers. We use this entropy as the group-level self-uncertainty target:
\[
t(G_x)
=
H_{\mathrm{SAGE}}(G_x).
\]

\subsection{Task-Specific Answer Equivalence}

The answer-equivalence kernel adapts SAGE across answer spaces by preserving task-critical distinctions that generic embeddings may obscure.

\noindent\textbf{Categorical answers.}
For multiple-choice tasks, option labels define equivalence:
\[
K_{\mathrm{ans}}^{\mathrm{bin}}(a_i,a_j)
=
\mathbb{I}[a_i=a_j].
\]
This prevents mutually exclusive options from being merged by embedding similarity.

\noindent\textbf{Numeric answers.}
For numeric reasoning, final values are normalized before comparison:
\[
\begin{aligned}
K_{\mathrm{ans}}^{\mathrm{bin}}(a_i,a_j)
=
\mathbb{I}\big[
&\operatorname{NormNum}(a_i)
\\
&=
\operatorname{NormNum}(a_j)
\big].
\end{aligned}
\]
This groups equivalent numeric forms while separating distinct values.

\noindent\textbf{Symbolic answers.}
For symbolic mathematics, final expressions are normalized and checked for equivalence:
\[
K_{\mathrm{ans}}^{\mathrm{bin}}(a_i,a_j)
=
\mathbb{I}[
\operatorname{Verify}(a_i,a_j)=1
].
\]
This preserves mathematical equivalence beyond surface form.

\noindent\textbf{Free-form answers.}
For open-ended factual QA, equivalence can be defined by normalized aliases or bidirectional entailment, grouping paraphrases while separating incompatible facts.

\section{Experiments}

\subsection{Experimental Setup}

\paragraph{Benchmarks.}
We evaluate on TriviaQA \cite{Joshi2017TriviaQAAL}, MATH-500 \cite{Hendrycks2021MeasuringMP}, and MMLU-Pro \cite{Wang2024MMLUProAM}. These benchmarks cover three representative answer structures: free-form factual answers, numeric or symbolic mathematical answers, and discrete multiple-choice options.

\paragraph{Compared methods.}
We compare GUPO with representative verbal uncertainty estimation and alignment baselines: Direct Verbalized prompting \cite{Xiong2023CanLE}, CoT prompting \cite{Cole2023SelectivelyAA}, Direct Confidence Alignment (DCA) \cite{Zhang2025DirectCA}, CSFT \cite{Jang2025VerbalizedCT}, and LoVeC-DPO \cite{Zhang2025ReinforcementLF}. GUPO uses SAGE as the group-level self-uncertainty target. Confidence-based outputs are converted into uncertainty scores for comparison. Further details on the baselines are provided in Appendix~\ref{app:baselines}.

\paragraph{Metrics.}
We evaluate calibration quality and uncertainty ranking. Brier score and Expected Calibration Error (ECE) measure whether the model's expressed uncertainty is calibrated to empirical reliability. Spearman correlation evaluates whether the uncertainty scores correctly rank examples by reliability. We also report @80, @60, and @50, following the standard high-confidence subset accuracy protocol. In our uncertainty framing, these metrics evaluate whether predictions assigned lower uncertainty are more likely to be correct. Higher values are better for Spearman, @80, @60, and @50, while lower values are better for Brier score and ECE. Further details on the conversion between uncertainty and confidence-based metrics are provided in Appendix~\ref{app:metrics}.

\subsection{Main Results}

We evaluate verbal uncertainty using both calibration and decision-oriented metrics. Brier score and ECE measure whether uncertainty has the correct numerical scale, but our main focus is ranking and selective reliability. Spearman correlation measures whether lower uncertainty corresponds to higher correctness, while @80, @60, and @50 evaluate whether low-uncertainty predictions form more accurate subsets. These metrics directly test whether verbal uncertainty can guide downstream decisions such as trusting, verifying, or abstaining. This distinction is important because a model may appear numerically calibrated while still failing to identify which individual predictions are reliable. In practical deployments, uncertainty is useful only if it supports such instance-level decisions. Therefore, improvements in ranking and selective accuracy provide stronger evidence that uncertainty expressions are behaviorally meaningful.

Table~\ref{tab:calibration_results_all} reports results on MMLU-Pro, MATH-500, and TriviaQA. Overall, \textbf{GUPO + SAGE} achieves the strongest profile. It obtains the best result on every metric for MMLU-Pro, the best Brier score and ECE on MATH-500 while remaining competitive on ranking metrics, and the clearest gains on TriviaQA across calibration, Spearman correlation, and high-confidence accuracy. These results show that GUPO + SAGE improves both the scale and, more importantly, the ordering of verbal uncertainty: predictions assigned lower uncertainty are more likely to be correct.

\begin{figure}[t]
    \centering
    \includegraphics[width=0.95\linewidth]{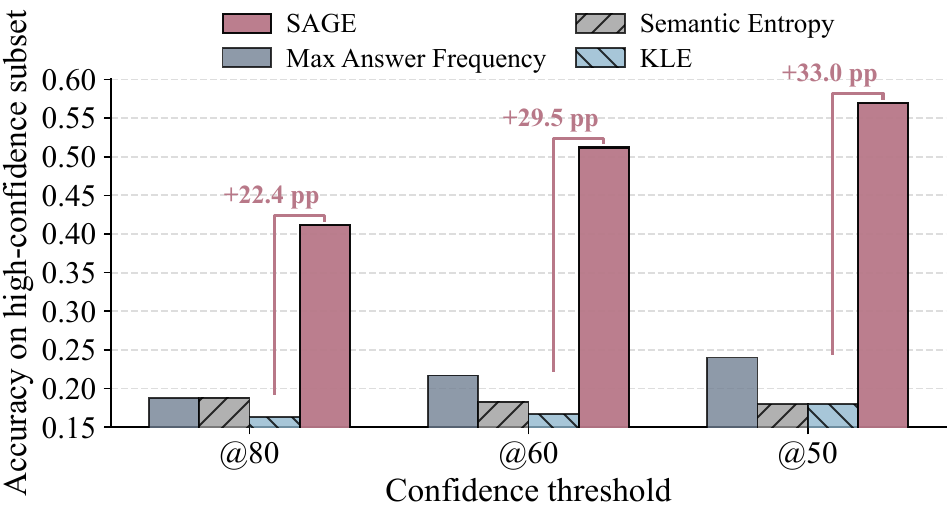}
   \caption{
    Ablation over group-level uncertainty targets on MMLU-Pro.
    SAGE achieves the best low-uncertainty subset accuracy across all thresholds, improving over the strongest baseline by \(+22.4\), \(+29.5\), and \(+33.0\) percentage points.
    These results show that SAGE better identifies reliable low-uncertainty predictions.
    }
    \label{fig:math500_target_ablation}
\end{figure}

\subsection{Ablation Study}
\begin{table*}[t]
\centering
\small
\renewcommand{\arraystretch}{1.14}
\setlength{\tabcolsep}{5.5pt}
\definecolor{oursblue}{RGB}{226,235,248}
\definecolor{deltagrey}{RGB}{240,240,240}

\begin{tabular}{llccccccc}
\toprule
\textbf{Model}
& \textbf{Setting}
& \textbf{Brier} $\downarrow$
& \textbf{ECE} $\downarrow$
& \textbf{Spear.} $\uparrow$
& \textbf{@80} $\uparrow$
& \textbf{@60} $\uparrow$
& \textbf{@50} $\uparrow$
& \textbf{$\Delta_\mu$} $\downarrow$ \\
\midrule

\multirow{3}{*}{Qwen2.5-1.5B-Instruct}
& Prompt & 0.359 & 0.546 & 0.246 & 0.562 & 0.617 & 0.660 & 0.526 \\
\cmidrule(lr){2-9}
& \cellcolor{oursblue}\textbf{GUPO+SAGE} 
& \cellcolor{oursblue}\textbf{0.020} 
& \cellcolor{oursblue}\textbf{0.024} 
& \cellcolor{oursblue}\textbf{0.463} 
& \cellcolor{oursblue}\textbf{0.613} 
& \cellcolor{oursblue}\textbf{0.717} 
& \cellcolor{oursblue}\textbf{0.740} 
& \cellcolor{oursblue}\textbf{0.021} \\
& \cellcolor{deltagrey}$\Delta$ vs. Prompt 
& \cellcolor{deltagrey}$-$0.339 
& \cellcolor{deltagrey}$-$0.522 
& \cellcolor{deltagrey}+0.217 
& \cellcolor{deltagrey}+0.051 
& \cellcolor{deltagrey}+0.100 
& \cellcolor{deltagrey}+0.080 
& \cellcolor{deltagrey}$-$0.505 \\
\midrule

\multirow{3}{*}{Qwen2.5-3B-Instruct}
& Prompt & 0.401 & 0.619 & -0.036 & 0.613 & 0.633 & 0.620 & 0.619 \\
\cmidrule(lr){2-9}
& \cellcolor{oursblue}\textbf{GUPO+SAGE} 
& \cellcolor{oursblue}\textbf{0.019} 
& \cellcolor{oursblue}\textbf{0.063} 
& \cellcolor{oursblue}\textbf{0.440} 
& \cellcolor{oursblue}\textbf{0.625} 
& \cellcolor{oursblue}\textbf{0.700} 
& \cellcolor{oursblue}\textbf{0.680} 
& \cellcolor{oursblue}\textbf{0.062} \\
& \cellcolor{deltagrey}$\Delta$ vs. Prompt 
& \cellcolor{deltagrey}$-$0.382 
& \cellcolor{deltagrey}$-$0.556 
& \cellcolor{deltagrey}+0.476 
& \cellcolor{deltagrey}+0.012 
& \cellcolor{deltagrey}+0.067 
& \cellcolor{deltagrey}+0.060 
& \cellcolor{deltagrey}$-$0.557 \\
\midrule

\multirow{3}{*}{Qwen2.5-7B-Instruct}
& Prompt & 0.389 & 0.602 & 0.381 & \textbf{0.588} & 0.583 & 0.660 & 0.602 \\
\cmidrule(lr){2-9}
& \cellcolor{oursblue}\textbf{GUPO+SAGE} 
& \cellcolor{oursblue}\textbf{0.024} 
& \cellcolor{oursblue}\textbf{0.035} 
& \cellcolor{oursblue}\textbf{0.485} 
& \cellcolor{oursblue}0.562 
& \cellcolor{oursblue}\textbf{0.650} 
& \cellcolor{oursblue}\textbf{0.700} 
& \cellcolor{oursblue}\textbf{0.035} \\
& \cellcolor{deltagrey}$\Delta$ vs. Prompt 
& \cellcolor{deltagrey}$-$0.365 
& \cellcolor{deltagrey}$-$0.567 
& \cellcolor{deltagrey}+0.104 
& \cellcolor{deltagrey}$-$0.026 
& \cellcolor{deltagrey}+0.067 
& \cellcolor{deltagrey}+0.040 
& \cellcolor{deltagrey}$-$0.567 \\

\bottomrule
\end{tabular}
\caption{
Calibration and uncertainty-alignment results across model sizes.
Prompt denotes direct uncertainty elicitation without training, while \textbf{GUPO+SAGE} denotes uncertainty-channel training with the proposed semantic-answer guided group-level target.
The mean-gap metric measures the absolute difference between the mean expressed uncertainty and the mean group-level uncertainty target.
High-confidence subset accuracy is computed after converting expressed uncertainty into confidence; in our uncertainty framing, it evaluates whether lower-uncertainty predictions are more reliable.
The delta rows report changes relative to Prompt; negative values are better for Brier score, ECE, and the mean-gap metric, while positive values are better for Spearman correlation and high-confidence subset accuracy.
}
\label{tab:calibration_results_by_size}
\end{table*}

\begin{figure}[t]
    \centering
    \includegraphics[width=\linewidth]{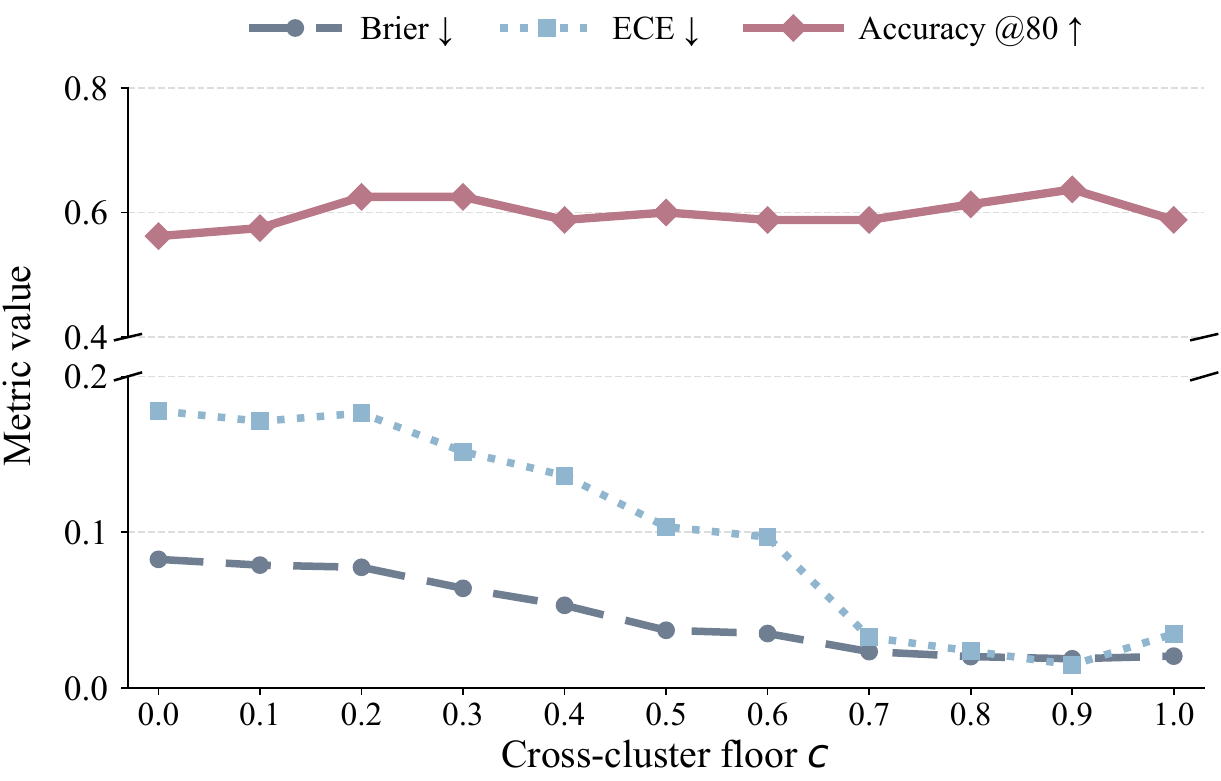}
    \caption{
    Ablation over the SAGE cross-cluster floor \(c\) on MATH-500.
    Increasing \(c\) improves calibration and low-uncertainty accuracy over the hard variant, while very large \(c\) moves the target closer to standard KLE.
    The results show that soft cross-cluster weighting provides a more effective training signal for SAGE.
    }
    \label{fig:sage_cfloor_ablation}
\end{figure}

\paragraph{Uncertainty target.}
We first compare four targets for constructing the calibration preference distribution, while keeping the GUPO training framework fixed: \textbf{Max answer frequency}, which uses the most frequent extracted answer and converts answer stability into an uncertainty target; \textbf{Semantic Entropy}, which computes uncertainty over meaning clusters; \textbf{KLE}, which estimates entropy in a continuous semantic embedding space; and \textbf{SAGE}, which constructs an answer-conditioned uncertainty geometry for response groups. This ablation isolates the central question of our method: whether a calibration target can simultaneously preserve answer-level disagreement, provide smooth reward variation, and avoid target-scale collapse. Frequency-based targets are answer-faithful but coarse; Semantic Entropy captures meaning equivalence but remains limited by hard cluster assignments; and KLE provides smoothness but can blur task-critical answer distinctions. In contrast, \textbf{SAGE is designed to satisfy all three requirements simultaneously}: answer faithfulness, reward smoothness, and scale preservation, making it a more effective target for converting group-level uncertainty into supervision for calibrated verbal uncertainty.

\paragraph{Effect of the cross-cluster floor.}
Figure~\ref{fig:sage_cfloor_ablation} ablates the soft cross-cluster floor \(c\) in SAGE. When \(c=0\), the answer kernel becomes hard, preserving answer distinctions but potentially compressing the target scale. Increasing \(c\) retains some linguistic similarity across different answer clusters, producing a smoother and more usable training signal. Very large \(c\), however, moves the target toward standard KLE and weakens the answer-aware correction. Overall, SAGE benefits from soft, scale-preserving answer separation rather than hard clustering or purely linguistic similarity.

\paragraph{Transferability Across Model Scales.}
Table~\ref{tab:calibration_results_by_size} evaluates GUPO across Qwen2.5-1.5B-Instruct, Qwen2.5-3B-Instruct, and Qwen2.5-7B-Instruct. Compared with direct prompting, GUPO+SAGE consistently reduces Brier score and ECE, and generally improves Spearman correlation. These results show that the proposed uncertainty-alignment framework transfers across model scales rather than relying on a single backbone.

Detailed ablation results and additional analyses are reported in Appendix~\ref{app:additional_experiments}.

\section{Conclusion}
We study verbal uncertainty alignment through behavioral uncertainty. Instead of treating uncertainty as a property of a single response, we define it by the stability of sampled outputs under the same prompt, framing calibration as a distributional problem. To derive a group-level target, we introduce SAGE, which combines continuous linguistic similarity with soft answer-equivalence structure. SAGE yields a smooth, answer-aware, scale-preserving uncertainty signal across factual, mathematical, and multiple-choice tasks. We further propose GUPO, which applies this signal to the verbal uncertainty channel rather than optimizing the full response. Experiments on TriviaQA, MATH-500, and MMLU-Pro show that GUPO improves alignment between expressed uncertainty and sampled model behavior, reducing overconfidence across task formats. These results demonstrate that group-level behavioral uncertainty can be distilled into single-response uncertainty expressions when both the target and optimization channel are properly designed.

\newpage
\section*{Limitations}

This work aligns verbal uncertainty with behavioral self-uncertainty rather than external correctness. The two are related but not identical: a model may be stable but wrong, or unstable while occasionally correct. Therefore, our method should be viewed as complementary to correctness-based calibration, retrieval-based verification, and external validation rather than as a replacement for them.

SAGE relies on answer extraction and task-specific equivalence rules to estimate group-level uncertainty. We use different normalization or equivalence procedures for factual, numeric, symbolic, and multiple-choice answers, but these procedures may still introduce noise, especially for open-ended tasks with ambiguous answer boundaries. In practice, this means that the quality of the uncertainty target depends partly on the reliability of the answer-equivalence module.

Finally, applying calibrated verbal uncertainty in high-stakes domains requires particular care. Better uncertainty expressions can help users recognize potentially unreliable outputs, but model predictions may still be affected by factual errors, incomplete evidence, hidden biases, or failures in the underlying reasoning process. A low-uncertainty response may still be incorrect, especially when the task involves domain-specific evidence or constraints. This is especially relevant in areas such as medicine, food safety, finance, and legal decision-making, where errors may have substantial real-world consequences. In these settings, model uncertainty can serve as a useful reference for risk awareness and verification, but final decisions should still be made by human experts and grounded in domain evidence, external validation, and appropriate institutional safeguards.



\newpage
\bibliography{custom}

@article{Savage2024LargeLM,
  title={Large language model uncertainty proxies: discrimination and calibration for medical diagnosis and treatment},
  author={Thomas Savage and John Wang and Robert J. Gallo and Abdessalem Boukil and Vishwesh Patel and Seyed Amir Ahmad Safavi-Naini and Ali Soroush and Jonathan H Chen},
  journal={Journal of the American Medical Informatics Association : JAMIA},
  year={2024},
  url={https://api.semanticscholar.org/CorpusID:273322711}
}

@article{Omar2025MultimodelAA,
  title={Multi-model assurance analysis showing large language models are highly vulnerable to adversarial hallucination attacks during clinical decision support},
  author={Mahmud Omar and Vera Sorin and Jeremy D. Collins and David L. Reich and Robert Freeman and Nicholas Gavin and Alexander W. Charney and Lisa Stump and Nicola Luigi Bragazzi and Girish N. Nadkarni and Eyal Klang},
  journal={Communications Medicine},
  year={2025},
  volume={5},
  url={https://api.semanticscholar.org/CorpusID:280415451}
}

@article{Meng2024TheAO,
  title={The application of large language models in medicine: A scoping review},
  author={Xiangbin Meng and Xiangyu Yan and Kuo Zhang and Da Liu and Xiaosheng Cui and Yaodong Yang and Muhan Zhang and Chunxia Cao and Jingjia Wang and Xuliang Wang and Jun Gao and Yuangengshuo Wang and Jiaming Ji and Zifeng Qiu and Muzi Li and Cheng Qian and Tianze Guo and Shuang Ma and Zeying Wang and Zexuan Guo and You-Lan Lei and Chunli Shao and Wen-yao Wang and Haojun Fan and Yifang Tang},
  journal={iScience},
  year={2024},
  volume={27},
  url={https://api.semanticscholar.org/CorpusID:269354751}
}

@article{Belkhouribchia2025LargeLM,
  title={Large language models in clinical nutrition: an overview of its applications, capabilities, limitations, and potential future prospects},
  author={Jamal Belkhouribchia and Joeri J. Pen},
  journal={Frontiers in Nutrition},
  year={2025},
  volume={12},
  url={https://api.semanticscholar.org/CorpusID:280703307}
}

@article{Farquhar2024DetectingHI,
  title={Detecting hallucinations in large language models using semantic entropy},
  author={Sebastian Farquhar and Jannik Kossen and Lorenz Kuhn and Yarin Gal},
  journal={Nature},
  year={2024},
  volume={630},
  pages={625 - 630},
  url={https://api.semanticscholar.org/CorpusID:270615909}
}

@article{Gupta2024LanguageMC,
  title={Language Model Cascades: Token-level uncertainty and beyond},
  author={Neha Gupta and Harikrishna Narasimhan and Wittawat Jitkrittum and Ankit Singh Rawat and Aditya Krishna Menon and Sanjiv Kumar},
  journal={ArXiv},
  year={2024},
  volume={abs/2404.10136},
  url={https://api.semanticscholar.org/CorpusID:269157057}
}

@article{Azaria2023TheIS,
  title={The Internal State of an LLM Knows When its Lying},
  author={Amos Azaria and Tom M. Mitchell},
  journal={ArXiv},
  year={2023},
  volume={abs/2304.13734},
  url={https://api.semanticscholar.org/CorpusID:258352729}
}

@article{Kuhn2023SemanticUL,
  title={Semantic Uncertainty: Linguistic Invariances for Uncertainty Estimation in Natural Language Generation},
  author={Lorenz Kuhn and Yarin Gal and Sebastian Farquhar},
  journal={ArXiv},
  year={2023},
  volume={abs/2302.09664},
  url={https://api.semanticscholar.org/CorpusID:257039062}
}

@article{Nikitin2024KernelLE,
  title={Kernel Language Entropy: Fine-grained Uncertainty Quantification for LLMs from Semantic Similarities},
  author={Alexander Nikitin and Jannik Kossen and Yarin Gal and Pekka Marttinen},
  journal={ArXiv},
  year={2024},
  volume={abs/2405.20003},
  url={https://api.semanticscholar.org/CorpusID:270123445}
}

@article{Xiong2023CanLE,
  title={Can LLMs Express Their Uncertainty? An Empirical Evaluation of Confidence Elicitation in LLMs},
  author={Miao Xiong and Zhiyuan Hu and Xinyang Lu and Yifei Li and Jie Fu and Junxian He and Bryan Hooi},
  journal={ArXiv},
  year={2023},
  volume={abs/2306.13063},
  url={https://api.semanticscholar.org/CorpusID:259224389}
}

@article{Lin2022TeachingMT,
  title={Teaching Models to Express Their Uncertainty in Words},
  author={Stephanie C. Lin and Jacob Hilton and Owain Evans},
  journal={Trans. Mach. Learn. Res.},
  year={2022},
  volume={2022},
  url={https://api.semanticscholar.org/CorpusID:249191391}
}

@article{Kapoor2024LargeLM,
  title={Large Language Models Must Be Taught to Know What They Don't Know},
  author={Sanyam Kapoor and Nate Gruver and Manley Roberts and Katherine M. Collins and Arka Pal and Umang Bhatt and Adrian Weller and Samuel Dooley and Micah Goldblum and Andrew Gordon Wilson},
  journal={ArXiv},
  year={2024},
  volume={abs/2406.08391},
  url={https://api.semanticscholar.org/CorpusID:270392060}
}

@inproceedings{Geng2023ASO,
  title={A Survey of Confidence Estimation and Calibration in Large Language Models},
  author={Jiahui Geng and Fengyu Cai and Yuxia Wang and Heinz Koeppl and Preslav Nakov and Iryna Gurevych},
  booktitle={North American Chapter of the Association for Computational Linguistics},
  year={2023},
  url={https://api.semanticscholar.org/CorpusID:265157516}
}

@article{Cole2023SelectivelyAA,
  title={Selectively Answering Ambiguous Questions},
  author={Jeremy R. Cole and Michael J.Q. Zhang and Daniel Gillick and Julian Martin Eisenschlos and Bhuwan Dhingra and Jacob Eisenstein},
  journal={ArXiv},
  year={2023},
  volume={abs/2305.14613},
  url={https://api.semanticscholar.org/CorpusID:258866001}
}

@article{Jang2025VerbalizedCT,
  title={Verbalized Confidence Triggers Self-Verification: Emergent Behavior Without Explicit Reasoning Supervision},
  author={Chaeyun Jang and Moonseok Choi and Yegon Kim and Hyungi Lee and Juho Lee},
  journal={ArXiv},
  year={2025},
  volume={abs/2506.03723},
  url={https://api.semanticscholar.org/CorpusID:279155057}
}

@article{Zhang2025DirectCA,
  title={Direct Confidence Alignment: Aligning Verbalized Confidence with Internal Confidence In Large Language Models},
  author={Glenn Zhang and Treasure Mayowa and Jason Fan and Yicheng Fu and Aaron Sandoval and Sean O'Brien and Kevin Zhu},
  journal={ArXiv},
  year={2025},
  volume={abs/2512.11998},
  url={https://api.semanticscholar.org/CorpusID:283897067}
}

@article{Zhang2025ReinforcementLF,
  title={Reinforcement Learning for Better Verbalized Confidence in Long-Form Generation},
  author={Caiqi Zhang and Xiaochen Zhu and Chengzu Li and Nigel Collier and Andreas Vlachos},
  journal={ArXiv},
  year={2025},
  volume={abs/2505.23912},
  url={https://api.semanticscholar.org/CorpusID:279068470}
}

@article{Joshi2017TriviaQAAL,
  title={TriviaQA: A Large Scale Distantly Supervised Challenge Dataset for Reading Comprehension},
  author={Mandar Joshi and Eunsol Choi and Daniel S. Weld and Luke Zettlemoyer},
  journal={ArXiv},
  year={2017},
  volume={abs/1705.03551},
  url={https://api.semanticscholar.org/CorpusID:26501419}
}

@article{Hendrycks2021MeasuringMP,
  title={Measuring Mathematical Problem Solving With the MATH Dataset},
  author={Dan Hendrycks and Collin Burns and Saurav Kadavath and Akul Arora and Steven Basart and Eric Tang and Dawn Xiaodong Song and Jacob Steinhardt},
  journal={ArXiv},
  year={2021},
  volume={abs/2103.03874},
  url={https://api.semanticscholar.org/CorpusID:232134851}
}

@article{Wang2024MMLUProAM,
  title={MMLU-Pro: A More Robust and Challenging Multi-Task Language Understanding Benchmark},
  author={Yubo Wang and Xueguang Ma and Ge Zhang and Yuansheng Ni and Abhranil Chandra and Shiguang Guo and Weiming Ren and Aaran Arulraj and Xuan He and Ziyan Jiang and Tianle Li and Max W.F. Ku and Kai Wang and Alex Zhuang and Rongqi "Richard" Fan and Xiang Yue and Wenhu Chen},
  journal={ArXiv},
  year={2024},
  volume={abs/2406.01574},
  url={https://api.semanticscholar.org/CorpusID:270210486}
}

@article{Yang2024Qwen25TR,
  title={Qwen2.5 Technical Report},
  author={Qwen An Yang and Baosong Yang and Beichen Zhang and Binyuan Hui and Bo Zheng and Bowen Yu and Chengyuan Li and Dayiheng Liu and Fei Huang and Guanting Dong and Haoran Wei and Huan Lin and Jian Yang and Jianhong Tu and Jianwei Zhang and Jianxin Yang and Jiaxin Yang and Jingren Zhou and Junyang Lin and Kai Dang and Keming Lu and Keqin Bao and Kexin Yang and Le Yu and Mei Li and Mingfeng Xue and Pei Zhang and Qin Zhu and Rui Men and Runji Lin and Tianhao Li and Tingyu Xia and Xingzhang Ren and Xuancheng Ren and Yang Fan and Yang Su and Yi-Chao Zhang and Yunyang Wan and Yuqi Liu and Zeyu Cui and Zhenru Zhang and Zihan Qiu and Shanghaoran Quan and Zekun Wang},
  journal={ArXiv},
  year={2024},
  volume={abs/2412.15115},
  url={https://api.semanticscholar.org/CorpusID:274859421}
}

@article{Chaudhry2024FinetuningLM,
  title={Finetuning Language Models to Emit Linguistic Expressions of Uncertainty},
  author={Arslan Chaudhry and Sridhar Thiagarajan and Dilan G{\"o}r{\"u}r},
  journal={ArXiv},
  year={2024},
  volume={abs/2409.12180},
  url={https://api.semanticscholar.org/CorpusID:272703963}
}

@inproceedings{Li2026ORCEOA,
  title={ORCE: Order-Aware Alignment of Verbalized Confidence in Large Language Models},
  author={Chen Li and Xiaolin Hu and Songzhu Zheng and Jiawei Zhou and Chaoran Chen},
  year={2026},
  url={https://api.semanticscholar.org/CorpusID:288259355}
}

@article{ye2025llms4all,
  title={Llms4all: A review of large language models across academic disciplines},
  author={Ye, Yanfang and Zhang, Zheyuan and Ma, Tianyi and Wang, Zehong and Li, Yiyang and Hou, Shifu and Sun, Weixiang and Shi, Kaiwen and Ma, Yijun and Song, Wei and others},
  journal={arXiv preprint arXiv:2509.19580},
  year={2025}
}

@inproceedings{chen2025clear,
  title={Clear: Towards contextual llm-empowered privacy policy analysis and risk generation for large language model applications},
  author={Chen, Chaoran and Zhou, Daodao and Ye, Yanfang and Li, Toby Jia-jun and Yao, Yaxing},
  booktitle={Proceedings of the 30th International Conference on Intelligent User Interfaces},
  pages={277--297},
  year={2025}
}

@inproceedings{chen2025obvious,
  title={The obvious invisible threat: Llm-powered gui agents’ vulnerability to fine-print injections},
  author={Chen, Chaoran and Zhang, Zhiping and Guo, Bingcan and Ma, Shang and Khalilov, Ibrahim and Gebreegziabher, Simret and Ye, Yanfang and Xiao, Ziang and Yao, Yaxing and Li, Tianshi and others},
  booktitle={Soups},
  year={2025},
  organization={The Twenty-First Symposium on Usable Privacy and Security (SOUPS)}
}

@inproceedings{li2024cheffusion,
  title={Cheffusion: Multimodal foundation model integrating recipe and food image generation},
  author={Li, Peiyu and Huang, Xiaobao and Tian, Yijun and Chawla, Nitesh V},
  booktitle={CIKM},
  year={2024}
}

@article{li2025adaptive,
  title={Adaptive Testing for LLM Evaluation: A Psychometric Alternative to Static Benchmarks},
  author={Li, Peiyu and Tang, Xiuxiu and Chen, Si and Cheng, Ying and Metoyer, Ronald and Hua, Ting and Chawla, Nitesh V},
  journal={arXiv},
  year={2025}
}

@article{li2025crochetbench,
  title={CrochetBench: Can Vision-Language Models Move from Describing to Doing in Crochet Domain?},
  author={Li, Peiyu and Huang, Xiaobao and Hua, Ting and Chawla, Nitesh V},
  journal={arXiv},
  year={2025}
}

@inproceedings{xiong2025deliberate,
  title={Deliberate reasoning in language models as structure-aware planning with an accurate world model},
  author={Xiong, Siheng and Payani, Ali and Yang, Yu’an and Fekri, Faramarz},
  booktitle={Proceedings of the 63rd Annual Meeting of the Association for Computational Linguistics (Volume 1: Long Papers)},
  pages={31900--31931},
  year={2025}
}

@inproceedings{xiong2025enhancing,
  title={Enhancing Language Model Reasoning with Structured Multi-Level Modeling},
  author={Xiong, Siheng and Payani, Ali and Fekri, Faramarz},
  booktitle={The Fourteenth International Conference on Learning Representations},
  year={2025}
}

@article{xiong2026scaling,
  title={Scaling Search-Augmented LLM Reasoning via Adaptive Information Control},
  author={Xiong, Siheng and Gungordu, Oguzhan and Johnson, Blair and Kerce, James C and Fekri, Faramarz},
  journal={arXiv preprint arXiv:2602.01672},
  year={2026}
}

@inproceedings{zhang2026mapro,
  title={MAPRO: Recasting Multi-Agent Prompt Optimization as Maximum a Posteriori Inference},
  author={Zhang, Zheyuan and Ge, Lin and Li, Hongjiang and Zhu, Weicheng and Zhang, Chuxu and Ye, Yanfang},
  booktitle={Findings of the Association for Computational Linguistics: EACL 2026},
  pages={4458--4480},
  year={2026}
}

@article{zhang2025agentrouter,
  title={AgentRouter: A Knowledge-Graph-Guided LLM Router for Collaborative Multi-Agent Question Answering},
  author={Zhang, Zheyuan and Shi, Kaiwen and Yuan, Zhengqing and Wang, Zehong and Ma, Tianyi and Murugesan, Keerthiram and Galassi, Vincent and Zhang, Chuxu and Ye, Yanfang},
  journal={arXiv preprint arXiv:2510.05445},
  year={2025}
}

@inproceedings{shi2026ng,
  title={NG-Router: Graph-Supervised Multi-Agent Collaboration for Nutrition Question Answering},
  author={Shi, Kaiwen and Zhang, Zheyuan and Yuan, Zhengqing and Murugesan, Keerthiram and Galassi, Vincent and Zhang, Chuxu and Ye, Yanfang},
  booktitle={Proceedings of the 19th Conference of the European Chapter of the Association for Computational Linguistics (Volume 1: Long Papers)},
  pages={7508--7527},
  year={2026}
}

@article{huang2026evolverouter,
  title={EvolveRouter: Co-Evolving Routing and Prompt for Multi-Agent Question Answering},
  author={Huang, Jiatan and Zhang, Zheyuan and Shi, Kaiwen and Ye, Yanfang and Zhang, Chuxu},
  journal={arXiv preprint arXiv:2604.05149},
  year={2026}
}

@article{huang2026glen,
  title={GLEN-Bench: A Graph-Language based Benchmark for Nutritional Health},
  author={Huang, Jiatan and Zhang, Zheyuan and Ma, Tianyi and Li, Mingchen and Zheng, Yaning and Ye, Yanfang and Zhang, Chuxu},
  journal={arXiv preprint arXiv:2601.18106},
  year={2026}
}

@article{bao2026drift,
  title={Drift-Bench: Diagnosing Cooperative Breakdowns in LLM Agents under Input Faults via Multi-Turn Interaction},
  author={Bao, Han and Zhang, Zheyuan and Jing, Pengcheng and Yuan, Zhengqing and Shi, Kaiwen and Ye, Yanfang},
  journal={arXiv preprint arXiv:2602.02455},
  year={2026}
}

@inproceedings{fadeeva2023lm,
  title={LM-polygraph: Uncertainty estimation for language models},
  author={Fadeeva, Ekaterina and Vashurin, Roman and Tsvigun, Akim and Vazhentsev, Artem and Petrakov, Sergey and Fedyanin, Kirill and Vasilev, Daniil and Goncharova, Elizaveta and Panchenko, Alexander and Panov, Maxim and others},
  booktitle={Proceedings of the 2023 Conference on Empirical Methods in Natural Language Processing: System Demonstrations},
  pages={446--461},
  year={2023}
}

@inproceedings{fadeeva2024fact,
  title={Fact-checking the output of large language models via token-level uncertainty quantification},
  author={Fadeeva, Ekaterina and Rubashevskii, Aleksandr and Shelmanov, Artem and Petrakov, Sergey and Li, Haonan and Mubarak, Hamdy and Tsymbalov, Evgenii and Kuzmin, Gleb and Panchenko, Alexander and Baldwin, Timothy and others},
  booktitle={Findings of the Association for Computational Linguistics: ACL 2024},
  pages={9367--9385},
  year={2024}
}

@article{hu2026entropy,
  title={Entropy-Gated Selective Policy Optimization: Token-Level Gradient Allocation for Hybrid Training of Large Language Models},
  author={Hu, Yuelin and Cheng, Zhengxue and Liu, Wei and Song, Li},
  journal={arXiv preprint arXiv:2602.03309},
  year={2026}
}

@article{xiong2023can,
  title={Can llms express their uncertainty? an empirical evaluation of confidence elicitation in llms},
  author={Xiong, Miao and Hu, Zhiyuan and Lu, Xinyang and Li, Yifei and Fu, Jie and He, Junxian and Hooi, Bryan},
  journal={arXiv preprint arXiv:2306.13063},
  year={2023}
}

@article{Manakul2023SelfCheckGPTZB,
  title={SelfCheckGPT: Zero-Resource Black-Box Hallucination Detection for Generative Large Language Models},
  author={Potsawee Manakul and Adian Liusie and Mark John Francis Gales},
  journal={ArXiv},
  year={2023},
  volume={abs/2303.08896},
  url={https://api.semanticscholar.org/CorpusID:257557820}
}

@article{Zhang2024LUQLU,
  title={LUQ: Long-text Uncertainty Quantification for LLMs},
  author={Caiqi Zhang and Fangyu Liu and Marco Basaldella and Nigel Collier},
  journal={ArXiv},
  year={2024},
  volume={abs/2403.20279},
  url={https://api.semanticscholar.org/CorpusID:268793903}
}

@article{Jiang2024GraphbasedUM,
  title={Graph-based Uncertainty Metrics for Long-form Language Model Outputs},
  author={Mingjian Jiang and Yangjun Ruan and Prasanna Sattigeri and Salim Roukos and Tatsunori B. Hashimoto},
  journal={ArXiv},
  year={2024},
  volume={abs/2410.20783},
  url={https://api.semanticscholar.org/CorpusID:273654396}
}

@article{nikitin2024kernel,
  title={Kernel language entropy: Fine-grained uncertainty quantification for llms from semantic similarities},
  author={Nikitin, Alexander and Kossen, Jannik and Gal, Yarin and Marttinen, Pekka},
  journal={Advances in Neural Information Processing Systems},
  volume={37},
  pages={8901--8929},
  year={2024}
}

@article{zhang2026semantic,
  title={Why Semantic Entropy Fails: Geometry-Aware and Calibrated Uncertainty for Policy Optimization},
  author={Zhang, Zheyuan and Shi, Kaiwen and Bao, Han and Wang, Zehong and Ma, Tianyi and Ye, Yanfang},
  journal={arXiv preprint arXiv:2605.21801},
  year={2026}
}

@article{ma2026autodata,
  title={Autodata: A multi-agent system for open web data collection},
  author={Ma, Tianyi and Qian, Yiyue and Zhang, Zheyuan and Wang, Zehong and Qian, Xiaoye and Bai, Feifan and Ding, Yifan and Luo, Xuwei and Zhang, Shinan and Murugesan, Keerthiram and others},
  journal={Advances in Neural Information Processing Systems},
  volume={38},
  pages={173416--173448},
  year={2026}
}

\newpage
\appendix


\section{Related Work}

LLMs have advanced rapidly in recent years \cite{ye2025llms4all, xiong2025deliberate, xiong2025enhancing, xiong2026scaling, li2024cheffusion, li2025adaptive, li2025crochetbench, ma2026autodata}. Building on this progress, LLM-driven agents have gained prominence for their ability to plan, interact, and solve complex tasks with limited human oversight \cite{zhang2026mapro, zhang2025agentrouter, shi2026ng, huang2026evolverouter, huang2026glen, bao2026drift}. A central practical challenge accompanying these successes is hallucination: models confidently producing incorrect or fabricated facts. This leads to the growing research of reliably quantifying and leveraging model uncertainty to faithfully represents model capabilities to recognize and mitigate hallucinations in trustworthy AI. 

A primary line of work estimates uncertainty directly from model predictions, where token-level entropy is used as a proxy for confidence. These methods interpret uncertainty as dispersion over next-token distributions and have been widely used in early studies \cite{fadeeva2023lm, fadeeva2024fact}. More recent approaches incorporate such signals into training, for example by using entropy to modulate gradient magnitude while preserving update direction \cite{hu2026entropy}. While effective for capturing local prediction uncertainty, these approaches remain fundamentally limited: token-level entropy reflects lexical ambiguity, but does not capture higher-level semantic structure across complete responses. As a result, it is often insufficient for reasoning-intensive tasks where correctness depends on global coherence rather than local token uncertainty.

To address this limitation, a growing body of work shifts to \emph{response-level uncertainty}, estimating confidence from agreement across multiple generated samples. Early approaches quantify agreement through exact-match frequency or consistency scores \cite{xiong2023can}, while later work extends this idea to semantic and factual consistency using external evaluators or LLM-based judgments \cite{Manakul2023SelfCheckGPTZB, Zhang2024LUQLU, Jiang2024GraphbasedUM}. Entropy-based formulations provide a more principled probabilistic interpretation by modeling distributions over semantic outcomes, including clustering-based semantic entropy \cite{Kuhn2023SemanticUL}, its discrete variant \cite{Farquhar2024DetectingHI}, and similarity-based extensions such as kernel entropy \cite{nikitin2024kernel} and methods beyond entropy \cite{zhang2026semantic}.

\section{Training Details}
\label{app:training_details}

\paragraph{Overview.}
All models are trained in two stages. 
First, we perform format supervised fine-tuning to make the model reliably produce task-specific answer and uncertainty formats. 
Second, starting from the format-SFT checkpoint, we train verbal uncertainty using group-level uncertainty targets. 
This second stage includes supervised calibration baselines and our uncertainty-channel preference training. 
All target variants share the same sampled response groups, data split, and evaluation protocol.

\begin{tcolorbox}[
colback=blue!3,
colframe=blue!45!black,
title=\textbf{Hardware and software},
fonttitle=\bfseries,
boxrule=0.6pt,
arc=2mm]
The main experiments use \texttt{Qwen/Qwen2.5-1.5B-Instruct} and are run on the PSC Bridges-2 cluster using the GPU-shared partition with a single NVIDIA V100-32GB GPU. 
For model-scale transfer experiments, \texttt{Qwen2.5-3B-Instruct} and \texttt{Qwen2.5-7B-Instruct} are trained on NVIDIA H100 GPUs. 
Training is implemented with TRL and Hugging Face \texttt{transformers}. 
Offline sampling and online rollouts use vLLM. 
Training and vLLM inference use fp16. 
For NLI-based semantic equivalence components, we use fp32 when required for attention compatibility. 
All experiments use seed 42 unless otherwise specified.
\end{tcolorbox}

\subsection{Training Model Selection}
We use \textbf{Qwen2.5-1.5B-Instruct} \cite{Yang2024Qwen25TR} as the primary training model. This choice is motivated by both methodological and practical considerations. Methodologically, verbal uncertainty calibration should be evaluated on a model that still exhibits visible uncertainty and non-trivial instability under repeated sampling; overly strong models may saturate on some benchmarks and make calibration improvements harder to diagnose. Practically, the 1.5B model enables efficient group rollout collection and preference optimization, which is important because GUPO requires multiple sampled responses for each prompt to estimate the SAGE target. Therefore, Qwen2.5-1.5B-Instruct provides a suitable testbed for studying whether SAGE can produce informative group-level supervision for uncertainty alignment.

To verify that the observed gains are not specific to a single model scale, we further evaluate the same training framework on \textbf{Qwen2.5-3B-Instruct} and \textbf{Qwen2.5-7B-Instruct}. As shown in Table~\ref{tab:calibration_results_by_size}, GUPO+SAGE consistently improves calibration across model sizes. In particular, it substantially reduces Brier score, ECE, and the mean uncertainty-target gap \(\Delta_\mu\), while improving Spearman correlation and high-confidence subset accuracy in most settings. These results suggest that the benefit of SAGE is not tied to the 1.5B model alone, but transfers across different model capacities.

\subsection{Metric Details}
\label{app:metrics}

Our model produces verbal uncertainty scores, while some standard calibration metrics are conventionally defined over confidence scores. For compatibility with prior work, we convert expressed uncertainty into the corresponding confidence score before computing Brier score, ECE, and threshold-based accuracy. Thus, high-confidence subset accuracy also has a direct uncertainty interpretation: examples above a confidence threshold are equivalently examples below the corresponding uncertainty threshold.

Specifically, @80, @60, and @50 report the accuracy of examples whose converted confidence score exceeds each threshold. In the uncertainty view, these metrics measure whether examples assigned lower uncertainty are more reliable. Therefore, higher @80, @60, and @50 indicate that the model's low-uncertainty predictions are more likely to be correct.

\subsection{Baseline Details}
\label{app:baselines}

We compare GUPO with representative baselines that cover prompting-based uncertainty elicitation, post-hoc calibration, supervised calibration training, and preference-based verbal uncertainty alignment.

\paragraph{Direct Verbalized.}
Direct Verbalized prompting asks the model to produce an answer together with an explicit verbal uncertainty statement. This baseline evaluates the model's native ability to express uncertainty without additional calibration training.

\paragraph{Chain-of-Thought Prompting.}
The CoT baseline asks the model to generate intermediate reasoning before producing the final answer and uncertainty expression. This tests whether adding reasoning improves the reliability of verbal uncertainty by making the answer process more explicit.

\paragraph{Direct Confidence Alignment.}
Direct Confidence Alignment (DCA) is a post-hoc calibration baseline that adjusts verbal confidence scores after generation. Since DCA operates on confidence values, we convert its calibrated confidence scores into uncertainty scores when comparing uncertainty alignment.

\paragraph{CSFT.}
CSFT uses supervised fine-tuning to train the model to produce calibrated verbal uncertainty expressions. Unlike GUPO, which uses group-level uncertainty targets and applies supervision to the uncertainty channel, CSFT follows a supervised response-level training objective.

\paragraph{LoVeC-DPO.}
LoVeC-DPO is a preference-based verbal calibration baseline. It improves verbal confidence expression through preference optimization, and we convert its confidence outputs into uncertainty scores for comparison. We include it to compare GUPO with an existing preference-based alignment approach.

\paragraph{GUPO with Alternative Targets.}
In addition to external baselines, we also evaluate GUPO with different group-level targets, including Semantic Entropy and Kernel Language Entropy. These variants use the same GUPO training framework as SAGE, but differ in how the group-level uncertainty target is constructed. This comparison isolates the contribution of SAGE from the optimization framework itself.

\paragraph{Dataset splits.}
We evaluate on three benchmarks with different answer structures. 
MATH-500 tests mathematical reasoning with numeric or symbolic final answers. 
TriviaQA tests factual question answering with short free-form textual answers. 
MMLU-Pro tests multiple-choice reasoning with discrete option labels. 
We use fixed train and validation splits for all calibration experiments. 
MATH-500 contains 500 prompts, split into 400 training prompts and 100 validation prompts. 
TriviaQA uses 2{,}000 prompts, split into 1{,}600 training prompts and 400 validation prompts. 
MMLU-Pro uses 4{,}000 prompts, split into 3{,}200 training prompts and 800 validation prompts. 
The same split is used across target variants and training methods.

\begin{tcolorbox}[
colback=green!3,
colframe=green!45!black,
title=\textbf{Offline response sampling for target construction},
fonttitle=\bfseries,
boxrule=0.6pt,
arc=2mm]
For each prompt \(x\), we sample a group
\[
G_x=\{y_1,\ldots,y_K\}, 
\qquad 
y_i\sim \pi_{\theta_0}(\cdot\mid x),
\]
where \(\theta_0\) is the base model and \(K=20\). 
All offline sampling uses vLLM with temperature \(0.7\) and top-\(p=0.9\). 
For MATH-500, the model produces chain-of-thought reasoning and a final \(\backslash\texttt{boxed}\{\cdot\}\) answer, with \(\texttt{max\_new\_tokens}=512\) and \(\texttt{max\_model\_len}=1536\). 
For TriviaQA, the model produces a short answer in an explicit \texttt{Answer:}/\texttt{Uncertainty:} format, with \(\texttt{max\_new\_tokens}=64\) and \(\texttt{max\_model\_len}=512\). 
For MMLU-Pro, the model produces chain-of-thought reasoning and ends with \(\texttt{Answer: <letter>}\), with \(\texttt{max\_new\_tokens}=320\) and \(\texttt{max\_model\_len}=2560\).
\end{tcolorbox}

\paragraph{Answer extraction and equivalence.}
For each sampled response \(y_i\), we extract the answer-bearing content \(z_i\), the final answer \(a_i\), and, when present, the verbal uncertainty expression \(v_i\). 
For MATH-500, we extract the final boxed answer and apply numeric or symbolic equivalence checking when possible. 
For TriviaQA, we normalize short answers by lowercasing, removing punctuation and articles, and using alias or semantic equivalence when available. 
For MMLU-Pro, we extract the final option label. 
These extracted answers are used for maximum answer frequency, semantic clustering, and the answer-equivalence kernel in SAGE.

\paragraph{Group-level target construction.}
Each target variant maps the response group \(G_x\) to a scalar uncertainty target
\[
t(G_x)\in[0,1],
\]
where larger values indicate greater uncertainty in the sampled response group. 
Maximum answer frequency is first computed as
\[
s_{\mathrm{MAF}}(G_x)
=
\max_a
\frac{1}{K}
\sum_{i=1}^{K}
\mathbb{I}[a_i=a].
\]
Since maximum answer frequency measures answer stability, its uncertainty version is given by
\[
t_{\mathrm{MAF}}(G_x)=1-s_{\mathrm{MAF}}(G_x).
\]
Semantic Entropy clusters responses into meaning-equivalence classes and computes uncertainty over the empirical cluster distribution. 
KLE estimates response dispersion in a continuous embedding space. 
SAGE combines linguistic similarity with answer-equivalence geometry. 
Specifically, SAGE uses
\[
\begin{aligned}
K^{\mathrm{soft}}_{ij}
&=
c+(1-c)K^{\mathrm{bin}}_{ij},
\\
K^{\mathrm{A}}_{ij}
&=
\left(K^{\mathrm{ling}}_{ij}\right)^\alpha
\left(K^{\mathrm{soft}}_{ij}\right)^\beta .
\end{aligned}
\]
Here \(K^{\mathrm{bin}}_{ij}=1\) when \(a_i\) and \(a_j\) are equivalent under the task-specific answer rule, and \(0\) otherwise. 
The cross-cluster floor \(c\in[0,1]\) controls how much similarity is retained between non-equivalent answer clusters. 
Unless otherwise specified, we use \(c=0.8\) in the main experiments and set \(\alpha=\beta=1\). 
The normalized entropy is used as the group-level uncertainty target:
\[
t(G_x)=H_{\mathrm{SAGE}}(G_x).
\]

\paragraph{Stage 1. Format supervised fine-tuning.}
The first training stage teaches the model to follow the output format required for answer and uncertainty extraction. 
This stage is not intended to calibrate uncertainty. 
Instead, it standardizes response structure across tasks. 
For MATH-500, the model learns to output reasoning followed by a boxed answer. 
For TriviaQA, it learns the explicit \texttt{Answer:}/\texttt{Uncertainty:} format. 
For MMLU-Pro, it learns to end with a final option label. 
All calibration experiments start from the corresponding format-SFT checkpoint, so that comparisons across targets and training methods are not confounded by formatting failures.

\paragraph{Stage 2. Calibration training.}
Starting from the format-SFT checkpoint, we train verbal uncertainty using group-level uncertainty targets. 
For supervised calibration baselines, the target \(t(G_x)\) is converted into a target uncertainty expression and optimized with supervised fine-tuning. 
For preference-based training, we use GUPO, which applies calibration supervision to the verbal uncertainty expression rather than to the full response. 
All target variants use the same offline response groups, and differ only in how \(t(G_x)\) is computed.

For each response \(y_i=(z_i,v_i)\), let \(u(v_i)\in[0,1]\) denote the extracted verbal uncertainty. 
Given \(t(G_x)\), we define the calibration score
\[
\begin{aligned}
r_i
&=
-\ell\big(u(v_i),t(G_x)\big),
\\
\ell\big(u(v_i),t(G_x)\big)
&=
\left(u(v_i)-t(G_x)\right)^2.
\end{aligned}
\]
The scores are normalized within the response group:
\[
p_i^{\mathrm{cal}}
=
\frac{\exp(r_i/T)}
{\sum_{j=1}^{K}\exp(r_j/T)}.
\]
The resulting preference signal is applied to the uncertainty channel \(v_i\) conditioned on the prompt and answer-bearing content \(z_i\). 
Conceptually, GUPO optimizes
\[
\begin{aligned}
\mathcal{L}_{\mathrm{GUPO}}
=
-\mathbb{E}_{x,G_x}
\Bigg[
&\sum_{i=1}^{K}
\operatorname{stopgrad}(p_i^{\mathrm{cal}})
\\
&\cdot
\log \pi_\theta(v_i\mid x,z_i)
\Bigg].
\end{aligned}
\]
In implementation, the calibration loss can be masked to the uncertainty-expression span.

\begin{tcolorbox}[
colback=orange!4,
colframe=orange!60!black,
title=\textbf{GUPO training hyperparameters},
fonttitle=\bfseries,
boxrule=0.6pt,
arc=2mm]
GUPO uses TRL with GUPO-style group preference training and vLLM rollouts. 
Unless otherwise specified, we use learning rate \(1\times 10^{-6}\), cosine scheduling, default linear warmup, and \(0.3\) epochs. 
The per-device batch size is 2, with gradient accumulation 8, giving an effective batch size of 16 prompts. 
Online training uses 8 rollouts per prompt. 
We use \(\texttt{max\_new\_tokens}=512\), maximum prompt length 384, KL coefficient \(0.1\), rollout temperature \(1.0\), and seed 42. 
All GUPO runs start from the format-SFT checkpoint.
\end{tcolorbox}

\paragraph{Evaluation protocol.}
At evaluation time, the trained model generates one response with verbal uncertainty for each validation prompt. 
We extract the final answer and numerical uncertainty. 
We report Brier score, Expected Calibration Error, Spearman correlation, and threshold-based accuracy at @80, @60, and @50. 
Threshold-based accuracy is computed after converting uncertainty into confidence, following the standard high-confidence subset accuracy protocol. 
Since all methods are evaluated on the same underlying sampled answer pool, raw answer accuracy is not the main comparison. 
The evaluation instead measures whether verbal uncertainty better ranks reliable examples and better matches the group-level uncertainty target.

\section{Additional Experiments}
\label{app:additional_experiments}

\subsection{Ablation over Group-Level Targets}
\begin{table*}[t]
\centering
\small
\renewcommand{\arraystretch}{1.15}
\setlength{\tabcolsep}{6.2pt}
\definecolor{sageblue}{RGB}{226,235,248}
\definecolor{headergrey}{RGB}{245,245,245}

\providecommand{\sg}{\cellcolor{sageblue}}

\begin{tabular}{llcccccc}
\toprule
\rowcolor{headergrey}
\textbf{Benchmark}
& \textbf{Target}
& \multicolumn{3}{c}{\textbf{Calibration}}
& \multicolumn{3}{c}{\textbf{High-Confidence Accuracy}} \\
\cmidrule(lr){3-5}
\cmidrule(lr){6-8}
\rowcolor{headergrey}
&
& \textbf{Brier} \(\downarrow\)
& \textbf{ECE} \(\downarrow\)
& \textbf{Spear.} \(\uparrow\)
& \textbf{@80} \(\uparrow\)
& \textbf{@60} \(\uparrow\)
& \textbf{@50} \(\uparrow\) \\
\midrule

\multirow{4}{*}{MMLU-Pro}
& Maximum Answer Frequency
& 0.184 & 0.301 & 0.210 & 0.188 & 0.217 & 0.240 \\
& Semantic Entropy
& 0.214 & 0.316 & 0.090 & 0.188 & 0.183 & 0.180 \\
& Kernel Language Entropy
& 0.068 & 0.205 & -0.010 & 0.163 & 0.167 & 0.180 \\
& \sg \textbf{SAGE}
& \sg \textbf{0.037} & \sg \textbf{0.113} & \sg \textbf{0.572}
& \sg \textbf{0.412} & \sg \textbf{0.512} & \sg \textbf{0.570} \\
\midrule

\multirow{4}{*}{MATH-500}
& Maximum Answer Frequency
& 0.155 & 0.252 & \textbf{0.515} & 0.588 & 0.683 & 0.700 \\
& Semantic Entropy
& 0.190 & 0.282 & 0.426 & 0.575 & 0.650 & 0.720 \\
& Kernel Language Entropy
& 0.0205 & 0.0348 & 0.451 & 0.588 & 0.700 & 0.720 \\
& \sg \textbf{SAGE}
& \sg \textbf{0.0202} & \sg \textbf{0.0238} & \sg 0.463
& \sg \textbf{0.613} & \sg \textbf{0.717} & \sg \textbf{0.740} \\
\midrule

\multirow{4}{*}{TriviaQA}
& Maximum Answer Frequency
& 0.158 & 0.252 & 0.296 & 0.502 & 0.521 & 0.540 \\
& Semantic Entropy
& 0.153 & 0.247 & 0.404 & 0.480 & 0.530 & 0.540 \\
& Kernel Language Entropy
& 0.123 & 0.227 & 0.268 & 0.490 & 0.480 & 0.520 \\
& \sg \textbf{SAGE}
& \sg \textbf{0.055} & \sg \textbf{0.031} & \sg \textbf{0.617}
& \sg \textbf{0.506} & \sg \textbf{0.583} & \sg \textbf{0.650} \\

\bottomrule
\end{tabular}
\caption{
Ablation over group-level uncertainty targets under the same GUPO training framework.
SAGE provides the strongest overall calibration signal across benchmarks, especially on calibration error, uncertainty ranking, and low-uncertainty subset accuracy.
}
\label{tab:gupo_target_ablation_full}
\end{table*}
\begin{table*}[t]
\centering
\small
\renewcommand{\arraystretch}{1.14}
\setlength{\tabcolsep}{4.6pt}
\definecolor{sageblue}{RGB}{226,235,248}
\definecolor{headergrey}{RGB}{245,245,245}

\begin{tabular}{lccccccccc}
\toprule
\rowcolor{headergrey}
\textbf{\(c\)}
& \multicolumn{2}{c}{\textbf{Target Distribution}}
& \multicolumn{1}{c}{\textbf{Verbal Unc.}}
& \multicolumn{3}{c}{\textbf{Calibration}}
& \multicolumn{3}{c}{\textbf{High-Confidence Accuracy}} \\
\cmidrule(lr){2-3}
\cmidrule(lr){4-4}
\cmidrule(lr){5-7}
\cmidrule(lr){8-10}
\rowcolor{headergrey}
& \textbf{Mean}
& \textbf{Std}
& \textbf{Mean \(u_v\)}
& \textbf{Brier} \(\downarrow\)
& \textbf{ECE} \(\downarrow\)
& \textbf{Spear.} \(\uparrow\)
& \textbf{@80} \(\uparrow\)
& \textbf{@60} \(\uparrow\)
& \textbf{@50} \(\uparrow\) \\
\midrule

0.0 & 0.1934 & 0.2393 & 0.000 & 0.0826 & 0.1779 & --    & 0.562 & 0.567 & 0.600 \\
0.1 & 0.1982 & 0.2364 & 0.013 & 0.0789 & 0.1712 & 0.138 & 0.575 & 0.583 & 0.600 \\
0.2 & 0.2102 & 0.2295 & 0.022 & 0.0775 & 0.1764 & 0.333 & 0.625 & 0.667 & 0.660 \\
0.3 & 0.2275 & 0.2202 & 0.065 & 0.0640 & 0.1517 & 0.304 & 0.625 & 0.667 & 0.640 \\
0.4 & 0.2492 & 0.2091 & 0.100 & 0.0531 & 0.1363 & 0.411 & 0.588 & 0.683 & 0.700 \\
0.5 & 0.2751 & 0.1970 & 0.164 & 0.0371 & 0.1035 & \textbf{0.524} & 0.600 & 0.683 & 0.700 \\
0.6 & 0.3049 & 0.1845 & 0.197 & 0.0350 & 0.0969 & 0.430 & 0.588 & \textbf{0.717} & \textbf{0.740} \\
0.7 & 0.3388 & 0.1724 & 0.306 & 0.0234 & 0.0326 & 0.429 & 0.588 & 0.700 & \textbf{0.740} \\
\rowcolor{sageblue}
0.8 & 0.3770 & 0.1620 & 0.347 & 0.0202 & 0.0238 & 0.463 & 0.613 & \textbf{0.717} & \textbf{0.740} \\
0.9 & 0.4201 & 0.1549 & 0.427 & \textbf{0.0188} & \textbf{0.0149} & 0.466 & \textbf{0.637} & 0.683 & 0.680 \\
1.0 & 0.4700 & 0.1544 & 0.497 & 0.0205 & 0.0348 & 0.451 & 0.588 & 0.700 & 0.720 \\

\bottomrule
\end{tabular}
\caption{
Sensitivity of SAGE to the soft cross-cluster floor \(c\) on MATH-500.
Here \(c=0\) corresponds to hard answer separation, while \(c=1\) reduces the target to the pure linguistic kernel.
The highlighted row denotes the default setting used in our main experiments.
}
\label{tab:cfloor_sensitivity_full}
\end{table*}

We further compare different group-level uncertainty targets under the same GUPO training framework. 
All variants use the same base model, sampled response groups, training protocol, and evaluation pipeline. 
The only difference is how the group-level target \(t(G_x)\) is computed. 
This ablation is designed to isolate the effect of target construction from the effect of the optimizer. 
In other words, all variants receive group rollouts and are trained with the same group-based optimization procedure, but the reward signal is induced by different uncertainty targets.

We compare four targets. 
\textbf{Max Answer Frequency} uses the share of the most frequent extracted answer in the group and converts answer stability into uncertainty. 
It is directly tied to answer agreement, but it compresses the group into a discrete count and cannot represent graded semantic or reasoning-level differences. 
\textbf{Semantic Entropy} clusters responses into meaning-equivalence classes and computes uncertainty over these clusters. 
It improves over exact answer counting by merging semantically equivalent generations, but its hard cluster assignments can produce discontinuous and low-resolution targets. 
\textbf{Hard SAGE} uses an answer-aware kernel with \(c=0\), which strictly separates different answer clusters. 
This maximizes answer separation, but can over-compress cross-answer similarity and collapse the target scale. 
\textbf{SAGE} uses a soft cross-cluster floor with \(c=0.8\), preserving answer-level distinctions while retaining enough cross-cluster similarity to maintain a usable reward scale.

Table~\ref{tab:gupo_target_ablation_full} reports the results. 
The comparison reveals several important patterns.

First, all non-SAGE targets show a mismatch between one desirable property and another. 
Semantic Entropy achieves moderate Spearman correlation and threshold accuracy, but its calibration errors remain high, with Brier score \(0.190\) and ECE \(0.282\). 
This indicates that semantic clustering can provide some ranking signal, but the resulting uncertainty values are poorly calibrated in absolute scale. 
Max Answer Frequency performs better than Semantic Entropy on Spearman correlation and high-confidence subset accuracy, reaching the best Spearman score among the non-SAGE variants. 
This is expected because answer frequency is closely tied to final-answer agreement. 
However, its Brier score and ECE remain substantially worse than SAGE, showing that answer frequency alone is too coarse to provide well-calibrated uncertainty values.

Second, the hard answer-aware variant confirms that answer awareness alone is not sufficient. 
Hard SAGE with \(c=0\) has much lower Brier and ECE than Semantic Entropy and Max Answer Frequency, but this improvement is misleading because the verbal uncertainty channel nearly collapses, with mean uncertainty only \(0.005\). 
The near-zero mean uncertainty also explains why Spearman correlation is weak. 
A hard block-diagonal answer kernel separates different answer clusters too aggressively, causing the target distribution to lose a usable scale for training. 
This supports our claim that the target must be answer-aware, but it must also be scale-preserving.

Third, SAGE with a soft cross-cluster floor provides the best overall trade-off. 
It achieves the lowest Brier score \(0.019\) and ECE \(0.022\), while also obtaining the best high-confidence subset accuracy across @50, @60, and @80. 
Its Spearman correlation \(0.513\) is comparable to Max Answer Frequency, but with far better calibration error. 
This is important because Spearman mainly measures ranking, while Brier and ECE measure whether the numerical uncertainty values are calibrated. 
SAGE therefore does not merely rank examples better; it also places uncertainty values on a more appropriate scale.

Overall, this ablation supports the central claim of our method. 
Group rollouts are useful only when the target extracted from the group provides a strong training signal. 
Max Answer Frequency is answer-faithful but too coarse. 
Semantic Entropy is more semantic but limited by hard clustering. 
Hard SAGE is answer-aware but collapses the uncertainty scale. 
Soft SAGE combines the useful properties of these targets: it preserves answer distinctions, remains smooth through the linguistic kernel, and avoids collapse through the cross-cluster floor. 
The resulting target gives GUPO a substantially more informative calibration signal.

\subsection{Sensitivity to the Cross-Cluster Floor}
\label{app:cfloor_sensitivity}

SAGE uses a soft cross-cluster floor \(c\) to control how much similarity is retained between responses with different extracted answers. 
When \(c=0\), the answer kernel becomes a hard block-diagonal kernel, which strongly separates different answer clusters but may collapse the target scale. 
When \(c=1\), the answer kernel becomes uninformative and SAGE reduces to the pure linguistic kernel. 
Intermediate values allow SAGE to preserve answer-level distinctions while retaining enough cross-cluster similarity to provide a usable training signal.

Table~\ref{tab:cfloor_sensitivity_full} reports the full sensitivity study on MATH-500.
As \(c\) increases, the target mean increases from \(0.1934\) to \(0.4700\), while the target standard deviation decreases from \(0.2393\) to \(0.1544\).
This indicates that the soft floor gradually relaxes hard answer separation and moves the target toward the smoother linguistic-kernel regime.
The hard setting \(c=0\) collapses the verbal uncertainty channel, producing \(\mathrm{mean}\ u_v=0.000\) and no valid Spearman correlation.
Soft floors substantially improve Brier score and ECE, with the strongest calibration performance around \(c=0.8\) to \(c=0.9\).
These results support the design choice that answer-aware geometry should be introduced softly rather than as hard answer separation.

\section{Mathematical Analysis of Target Identifiability and Reward Signal}
\subsection{Why Response-Level Supervision is Insufficient}

Let $x$ be a prompt and let the model induce a response distribution
\[
p_\theta(y\mid x).
\]
Each response $y_i$ contains answer-bearing content $z_i$, an extracted answer
\[
a_i = A(z_i),
\]
and a verbal uncertainty value $u_i \in [0,1]$. The goal of verbal uncertainty alignment is to make $u_i$ reflect the model's uncertainty over possible responses to the same prompt. Thus, the ideal uncertainty target is not a function of a single sampled response, but a functional of the response distribution:
\[
t^\star(x)
=
T\!\left(p_\theta(\cdot\mid x)\right).
\]
Equivalently, with a sampled group
\[
G_x=\{y_1,\ldots,y_K\},\qquad y_i\sim p_\theta(\cdot\mid x),
\]
we estimate
\[
t(G_x)\approx t^\star(x).
\]

\paragraph{Under-identification of a single response.}
A response-level method observes only one sample $y$ for each prompt. We show that this is insufficient because the same observed response can be generated from different underlying response distributions with different uncertainty.

Consider two answer distributions over $\{A,B,C,D\}$:
\[
p_1(a\mid x)=(0.90,0.04,0.03,0.03),
\]
\[
p_2(a\mid x)=(0.30,0.25,0.25,0.20).
\]
Suppose the sampled response in both cases gives answer $A$. From the perspective of a response-level objective, the observed instance is identical:
\[
(x,y,A).
\]
However, the correct uncertainty target should be very different. For example, under an entropy-based uncertainty target, the distribution \(p_1\) should receive lower uncertainty than \(p_2\), because its probability mass is more concentrated on one answer. Thus the same observed answer $A$ should receive low uncertainty under $p_1$ but much higher uncertainty under $p_2$.

This means any pointwise target of the form
\[
r(x,y,u)
\]
cannot identify the correct distributional uncertainty, because
\[
r(x,y,u;p_1)=r(x,y,u;p_2)
\]
whenever the method only observes the same $(x,y,u)$. The missing information is the distribution of alternative plausible outputs.

\paragraph{Consequence for SFT and DPO.}
SFT minimizes a response-level loss:
\[
\mathcal{L}_{\mathrm{SFT}}
=
-\log p_\theta(y,u^\star\mid x).
\]
This objective can imitate a provided uncertainty string $u^\star$, but it does not reveal whether $y$ came from a stable distribution or an unstable one.

Similarly, DPO-style methods compare two responses:
\[
\begin{aligned}
\mathcal{L}_{\mathrm{DPO}}
=
-\log \sigma
\Bigg(
\beta
\Big[
&\log\frac{\pi_\theta(y^+\mid x)}
{\pi_{\mathrm{ref}}(y^+\mid x)}
\\
&-
\log\frac{\pi_\theta(y^-\mid x)}
{\pi_{\mathrm{ref}}(y^-\mid x)}
\Big]
\Bigg).
\end{aligned}
\]
The preference pair $(y^+,y^-)$ can indicate which response is preferred, but it does not determine the full uncertainty distribution over possible generations. Many distributions can produce the same pair:
\[
(y^+,y^-)\subset \operatorname{supp}(p_\theta(\cdot\mid x)),
\]
while having different uncertainty targets
\[
T(p_\theta(\cdot\mid x)).
\]
Therefore, response-level supervision is under-identified for verbal uncertainty calibration. It can learn local uncertainty patterns, but it cannot determine whether the uncertainty should reflect a concentrated or dispersed response distribution.

\subsection{Why Existing Group-Level Targets Give Weak Rewards}

Group rollouts address the distributional challenge by exposing multiple samples:
\[
G_x=\{y_1,\ldots,y_K\}.
\]
However, the target computed from $G_x$ must also be useful for optimization. Suppose the training reward is
\[
R(u,G_x)
=
-\left(u-t(G_x)\right)^2,
\]
where \(t(G_x)\) is the group-level uncertainty target. If \(t(G_x)\) is coarse, discontinuous, or misaligned with answer disagreement, then the reward provides weak or misleading supervision.

\paragraph{Maximum answer frequency is coarse.}
The maximum answer frequency stability score is
\[
s_{\mathrm{MAF}}(G_x)
=
\max_a
\frac{1}{K}
\sum_{i=1}^K
\mathbb{I}[a_i=a].
\]
Its uncertainty version is
\[
t_{\mathrm{MAF}}(G_x)=1-s_{\mathrm{MAF}}(G_x).
\]
Let
\[
n_a=\sum_{i=1}^K \mathbb{I}[a_i=a].
\]
Then
\[
s_{\mathrm{MAF}}(G_x)=\frac{1}{K}\max_a n_a.
\]
Thus MAF depends only on the largest answer count and discards the rest of the group structure.

For example, with $K=10$,
\[
G_1=(6A,4B),
\qquad
G_2=(6A,1B,1C,1D,1E).
\]
Both groups receive the same stability score:
\[
s_{\mathrm{MAF}}(G_1)=s_{\mathrm{MAF}}(G_2)=0.6.
\]
But $G_2$ contains more diverse secondary alternatives. Hence the target collapses distinct uncertainty structures into the same reward value when it only uses the largest answer count.

Moreover, MAF can take only $K$ possible stability values:
\[
s_{\mathrm{MAF}}(G_x)
\in
\left\{
\frac{1}{K},\frac{2}{K},\ldots,1
\right\}.
\]
Therefore the reward landscape is discrete. If two groups have the same majority count, then for any uncertainty value \(u\),
\[
R(u,G_1)=R(u,G_2),
\]
even if their semantic variation or secondary answer disagreement is different. MAF is answer-faithful, but it is too coarse to provide fine-grained calibration supervision.

\paragraph{Semantic Entropy is discontinuous.}
Semantic Entropy clusters responses into meaning-equivalence classes. Let
\[
C(y_i)\in\{1,\ldots,M\}
\]
be the semantic cluster assignment, and let
\[
p_m
=
\frac{1}{K}
\sum_{i=1}^K
\mathbb{I}[C(y_i)=m].
\]
Then
\[
H_{\mathrm{SE}}(G_x)
=
-\sum_{m=1}^M p_m\log p_m.
\]
The target depends only on hard cluster counts. Therefore, if a response changes slightly but remains in the same cluster,
\[
C(y_i)=C(y_i')
\quad\Rightarrow\quad
H_{\mathrm{SE}}(G_x)=H_{\mathrm{SE}}(G_x').
\]
This means the reward is flat within each cluster:
\[
\Delta H_{\mathrm{SE}}=0.
\]

However, if a small change moves one response across a cluster boundary, then the empirical cluster distribution changes discontinuously. Suppose one response moves from cluster $u$ to cluster $v$. Then
\[
p_u' = p_u-\frac{1}{K},
\qquad
p_v' = p_v+\frac{1}{K}.
\]
The entropy change is
\[
\begin{aligned}
\Delta H_{\mathrm{SE}}
=
&-\left(p_u-\frac{1}{K}\right)
\log\left(p_u-\frac{1}{K}\right)
\\
&-\left(p_v+\frac{1}{K}\right)
\log\left(p_v+\frac{1}{K}\right)
\\
&+p_u\log p_u
+p_v\log p_v .
\end{aligned}
\]
This jump occurs because the cluster assignment changes, not because the response distribution has changed smoothly. Therefore SE produces a piecewise-constant reward:
\[
\text{within cluster: } \Delta H_{\mathrm{SE}}=0,
\]
\[
\text{across boundary: } \Delta H_{\mathrm{SE}}\neq 0.
\]
As a result, SE improves over exact answer counting for paraphrases, but its hard clustering gives weak credit assignment for fine-grained uncertainty calibration.

\paragraph{KLE is smooth but not answer-faithful.}
KLE replaces hard clusters with a continuous kernel. Let
\[
K_{ij}=k(e(y_i),e(y_j)),
\qquad K_{ij}\in[0,1],
\]
and normalize
\[
P=\frac{K}{\operatorname{tr}(K)}.
\]
The normalized von Neumann entropy is
\[
H_{\mathrm{KLE}}(G_x)
=
-\frac{1}{\log K}
\operatorname{tr}(P\log P).
\]
This target is smooth because small changes in embeddings produce small changes in pairwise similarities. However, smoothness alone is insufficient. KLE assumes that embedding similarity reflects answer compatibility:
\[
K_{ij}\text{ high}
\quad\Rightarrow\quad
a_i\equiv a_j.
\]
This assumption fails in structured answer spaces. For example, multiple-choice labels $A$, $B$, $C$, and $D$ may be close in embedding space even though they are mutually exclusive answers:
\[
A\not\equiv B\not\equiv C\not\equiv D.
\]

We can see the failure mathematically. Suppose a response group contains $K$ mutually incompatible answers, but generic embeddings assign high similarity to all pairs:
\[
K_{ii}=1,
\qquad
K_{ij}=r \quad (i\neq j),
\qquad r\approx 1.
\]
Then
\[
K=(1-r)I+r\mathbf{1}\mathbf{1}^{\top}.
\]
The eigenvalues of $K$ are
\[
\lambda_1=1+(K-1)r,
\qquad
\lambda_2=\cdots=\lambda_K=1-r.
\]
Since $\operatorname{tr}(K)=K$, the eigenvalues of $P$ are
\[
\begin{aligned}
\mu_1
&=
\frac{1+(K-1)r}{K},
\\
\mu_2=\cdots=\mu_K
&=
\frac{1-r}{K}.
\end{aligned}
\]
As $r\to 1$,
\[
\mu_1\to 1,
\qquad
\mu_2,\ldots,\mu_K\to 0.
\]
Therefore,
\[
H_{\mathrm{KLE}}(G_x)
=
-\frac{1}{\log K}
\sum_{i=1}^K \mu_i\log\mu_i
\to 0.
\]
So KLE predicts low uncertainty even when all final answers are incompatible. This shows that KLE is smooth, but its generic geometry can underestimate task-critical answer disagreement.

\subsection{Summary}

The above analysis shows that a useful group-level uncertainty target must satisfy three requirements:
\begin{enumerate}
    \item \textbf{Answer-faithfulness}: the target should reflect task-critical answer agreement and disagreement.
    \item \textbf{Reward smoothness}: the target should change continuously with meaningful variations in the response group.
    \item \textbf{Scale preservation}: the target should preserve a usable uncertainty range rather than collapsing to near-zero or near-one values.
\end{enumerate}
MAF is answer-faithful but coarse. SE handles paraphrases but produces flat and discontinuous rewards because of hard clustering. KLE is smooth but can collapse incompatible answers when generic embeddings place them close together. Therefore, repeated sampling is not sufficient by itself: the target must transform sampled responses into an answer-aware and optimization-friendly reward signal.

\section{Qualitative Case Study}

\begin{tcolorbox}[
    colback=blue!3,
    colframe=blue!45!black,
    title=\textbf{Case Study: When Semantic Similarity Hides Answer Disagreement},
    fonttitle=\bfseries,
    arc=1.5mm,
    boxrule=0.8pt,
    left=1.5mm,
    right=1.5mm,
    top=1mm,
    bottom=1mm
]
\textbf{Prompt.}
A multiple-choice question asks the model to select one option from \(\{A,B,C,D\}\).

\vspace{0.5em}
\textbf{Repeated samples from the model.}
\[
\begin{aligned}
G_x=\{&
(A,0.18),\ (A,0.21),\ (B,0.24),\\
&(C,0.26),\ (A,0.19),\ (B,0.23)
\}.
\end{aligned}
\]

\vspace{0.5em}
\textbf{Observed answer distribution.}
\[
\begin{aligned}
p(A)&=\frac{3}{6}, \qquad
p(B)=\frac{2}{6},\\
p(C)&=\frac{1}{6}, \qquad
p(D)=0.
\end{aligned}
\]
\vspace{0.5em}
\textbf{Interpretation.}
The model frequently expresses low uncertainty, but its sampled answers are not stable. The group contains mutually exclusive choices, so the appropriate uncertainty target should be higher than the expressed values suggest.
\end{tcolorbox}

\begin{center}
\small
\setlength{\tabcolsep}{4pt}
\renewcommand{\arraystretch}{1.15}
\begin{tabular}{lcc}
\toprule
\textbf{Target} 
& \makecell{\textbf{Behavior}\\\textbf{on this group}} 
& \makecell{\textbf{Limitation}\\\textbf{/ Effect}} \\
\midrule
MAF 
& \makecell{\(s_{\mathrm{MAF}}=3/6\)\\\(=0.50\)} 
& \makecell{answer-faithful\\but coarse} \\
KLE 
& \makecell{high similarity\\among option labels} 
& \makecell{may underestimate\\disagreement} \\
SAGE 
& \makecell{separates \(A,B,C\)\\as incompatible} 
& \makecell{answer-aware\\target} \\
\bottomrule
\end{tabular}
\end{center}

This example illustrates why generic semantic similarity can be misleading for uncertainty calibration. In a multiple-choice task, option labels such as \(A\), \(B\), and \(C\) are linguistically similar because they are short symbolic labels with nearly identical surface form. A generic embedding-based target may therefore assign high pairwise similarity to responses ending in different option labels. This can make the response group appear more coherent than it actually is, producing an overly low uncertainty target.

However, from the task perspective, \(A\), \(B\), and \(C\) are not nearby semantic variants. They are mutually exclusive decisions. A model that alternates among these choices should express higher uncertainty, even if the generated explanations have similar wording. SAGE corrects this mismatch by evaluating response groups under answer-aware equivalence. Responses with the same extracted option remain close, while responses with different option labels are separated in the answer-conditioned geometry. As a result, the group receives a higher and more appropriate uncertainty target, which gives GUPO a clearer supervision signal for reducing overconfident verbal expressions.

\section{Use of AI Assistants}

The authors used AI assistants only for language polishing and minor editing of the manuscript text. AI assistants were not used to generate research ideas, design the method, conduct experiments, produce results, or write technical claims. All content, including the proposed method, experimental design, analysis, and final manuscript, was reviewed, verified, and approved by the authors.

\end{document}